# Article

# Development and Comparative Analysis of Machine Learning Models for Hypoxemia Severity Triage in CBRNE Emergency Scenarios Using Physiological and Demographic Data from Medical-Grade Devices

Santino NANINI[1], Mariem ABID[2], Yassir MAMOUNI[3], Arnaud WIEDEMANN[1], Philippe JOUVET[1], and Stephane BOURASSA[1,4]

[1] **SADC-CDSS IA PEDIATRICS**, CHU Sainte-Justine, Montreal, Canada; santino.nanini@umontreal.com

[2] **Solutions Applicare AI Inc.**, Montreal, Canada; mariem.abid@applicare.ai

[3] **Université de Montréal**, Canada; yassir.mamouni@outlook.com

[4] **MEDINT CBRNE Group**, Montreal, Canada; sb@medintcbrne.com

[1] **SADC-CDSS IA PEDIATRICS**, CHU Sainte-Justine, Montreal, Canada; philippe.jouvet.med@ssss.gouv.qc.ca

[1] **SADC-CDSS IA PEDIATRICS**, CHU Sainte-Justine, Montreal, Canada; arnaud.wiedemann.med@ssss.gouv.qc.ca

* **Correspondence**: santino.nanini@umontreal.com
https://www.chusj-sip-ia.ca/

# Abstract

This paper presents the development of machine learning (ML) models to predict hypoxemia severity during emergency triage, especially in Chemical, Biological, Radiological, Nuclear, and Explosive (CBRNE) events, using physiological data from medical-grade sensors. Gradient Boosting Models (XGBoost, LightGBM, CatBoost) and sequential models (LSTM, GRU) were trained on physiological and demographic data from the MIMIC-III and IV datasets. A robust preprocessing pipeline addressed missing data, class imbalances, and incorporated synthetic data flagged with masks.

Gradient Boosting Models (GBMs) outperformed sequential models in terms of training speed, interpretability, and reliability, making them well-suited for real-time decision-making. While their performance was comparable to that of sequential models, the GBMs used score features from six physiological variables derived from the enhanced National Early Warning Score (NEWS) 2, which we termed NEWS2+. This approach significantly improved prediction accuracy.

While sequential models handled temporal data well, their performance gains didn't justify the higher computational cost. A 5-minute prediction window was chosen for timely intervention, with minute-level interpolations standardizing the data.

Feature importance analysis highlighted the significant role of mask and score features in enhancing both transparency and performance. Temporal dependencies proved to be less critical, as Gradient Boosting Models were able to capture key patterns effectively without relying on them.

This study highlights ML's potential to improve triage and reduce alarm fatigue. Future work will integrate data from multiple hospitals to enhance model generalizability across clinical settings.

# 1. Introduction

Rapid and accurate patient triage is essential during disaster situations, especially in Chemical, Biological, Radiological, Nuclear, and Explosive (CBRNE) events [1–4]. These incidents often lead to mass casualties and chaotic environments, overwhelming traditional triage systems. The VIMY Multi-System is a technologically enhanced response platform designed to improve casualty management in these complex scenarios. Initiated at the Grouping in AI acute care for the child (CHU Ste Justine, Montreal, Canada)[1*] as part of the VIMY research program, this system leverages artificial intelligence (AI) to manage casualties in disaster settings, including CBRNE events.

More specifically, the VIMY Multi-System [5–8] is a field-deployable intensive care unit that integrates AI, sensors, and decision-making algorithms to improve healthcare during disasters

---

[1]* https://www.chusj-sip-ia.ca/

(Figure 1). The core framework of the VIMY system incorporates an Electronic Casualty Card System (ECCS), functioning as a dashboard that integrates monitoring, scoring, and treatment data (Figure 2). Its primary aim is to equip frontline personnel with advanced tools to overcome the limitations of current triage and early warning systems (EWS), particularly in CBRNE scenarios, by harnessing machine learning to enhance patient care. This paper specifically focuses on the development of predictive algorithms that are fundamental to the ECCS, thereby contributing directly to the improvement of these critical systems within the VIMY project.

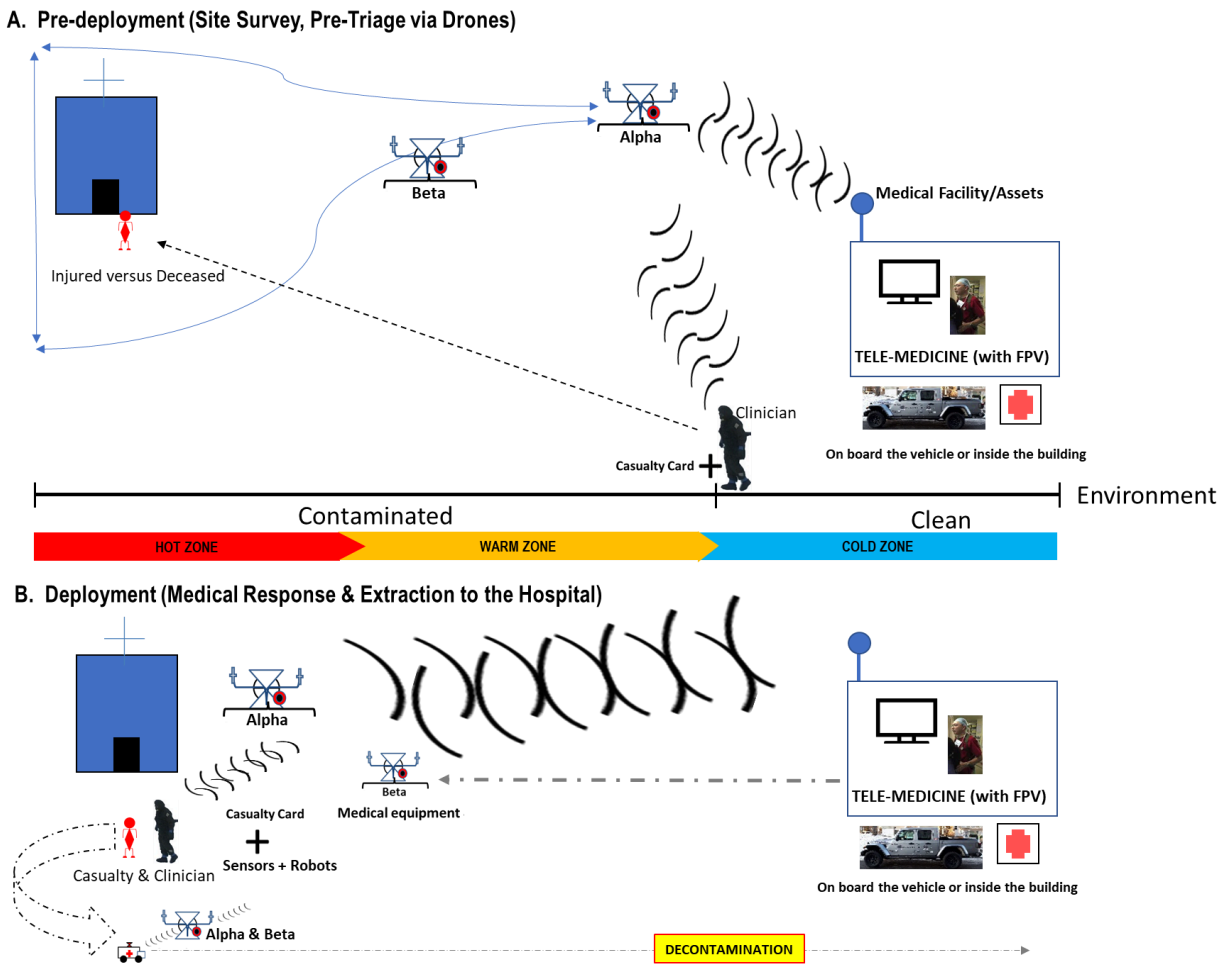

**Figure 1 An illustration that shows the VIMY Multisystem deployment** [8], in a glance, during a medical response in a contaminated environment. Our goal is to develop a proof of concept for AI models that will ensure effective triage in CBRNE environments and other pre-clinical settings.

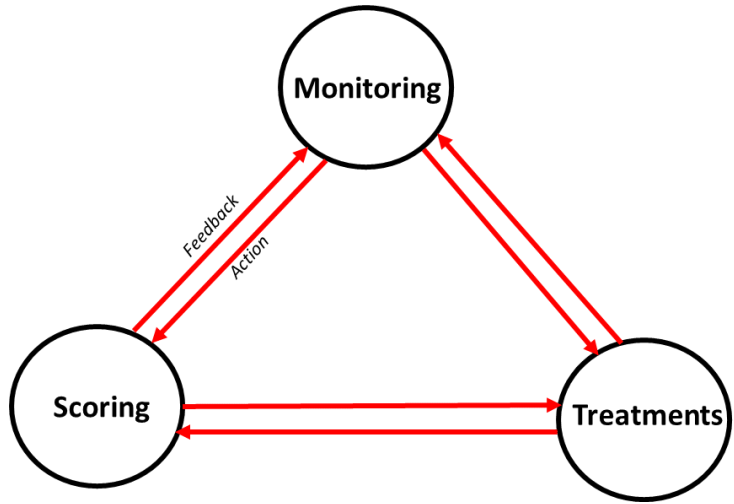

**Figure 2 The illustration represents a simulated hypoxemia condition, testing the monitoring-scoring-treatment nexus through an iterative process** [8]. 1. **Monitoring**: Continuous monitoring of vital signs provides real-time data on key physiological variables such as respiratory rate, temperature, blood pressure, heart rate, and SpO₂ (oxygen saturation). Both real and simulated data are analyzed to detect correlations and patterns in these digital biomarkers. This analysis helps in predicting the severity score of hypoxemia. 2. **Scoring**: Based on SpO₂ levels, a hypoxemia severity score is determined, ranging from 0 (best) to 3 (worst). This predicted severity score informs triage decisions, categorizing the patient's condition as Stat, Urgent, or Stable. 3. **Treatment Administration**: Depending on the severity score, appropriate treatments are administered. For instance, a patient with low oxygen saturation may receive oxygen therapy (with or without an oxygen mask) to stabilize their condition.

EWS are tools used in healthcare to assess the severity of a patient's condition by monitoring key physiological parameters, such as heart rate, blood pressure, respiratory rate, temperature, and level of consciousness. EWS helps healthcare providers identify patients at risk of deterioration, enabling timely interventions. By tracking vital signs and assigning scores based on their values, EWS aids in the early detection of patients who may need closer monitoring or medical intervention.

EWS such as NEWS2 [9–12], PWES [13–16], the Pediatric Vital Signs Normal Ranges from the Iowa Head and Neck Protocols [17,18], and the Modified Early Warning Score (MEWS) [19–21] have been developed to monitor patients' vital signs [11,22,23]. However, existing EWS do not predict the worsening of a patient's condition, which would be crucial for early intervention, especially in pre-clinical scenarios like CBRNE events [11,23,24], where multiple patients are injured simultaneously [24,25]. Thus, EWS are known to have limitations [23,24], even if they are very useful.

Machine learning (ML) offers a promising approach to enhance the predictive capabilities of EWS by analyzing complex physiological data [26,27]. In this study, we present the development of ML-based models using data from medical devices to predict hypoxemia severity, aiming to improve triage efficiency and reduce medical staff fatigue. Our models leverage physiological and demographic data (including age, which influences the medical interpretation of physiological constants), focusing on key vital signs such as respiratory rate,

SpO$_2$ (oxygen saturation) levels, heart rate, both systolic and diastolic blood pressure, and temperature.

Hypoxemia, characterized by low oxygen levels in the blood, is a common and critical condition in disaster settings. Accurate assessment of hypoxemia severity is essential for timely intervention. Current efforts to use ML to predict hypoxic events are happening in various settings and showcasing diverse methodologies. The studies differ considerably in patient populations, outcome definitions, predictive features, and ML algorithms, making it hard to generalize their conclusions. Consequently, comparing and evaluating these studies comprehensively is quite challenging [28]. Indeed, a systematic review compares [28] past efforts to predict hypoxic events in hospital settings using machine learning, focusing on methodologies, predictive performance, and the populations assessed. The authors identified 12 studies that predicted hypoxic events or hypoxia markers across various settings, including operating rooms, ICUs, and general care units. The machine learning models applied were based on both conventional ML and deep learning methods. Most studies defined their prediction endpoints using specific thresholds for blood oxygen measurements. Clinical variables included patient characteristics, vital signs, and laboratory data, with blood oxygen data (such as peripheral oxygen saturation, SpO$_2$ ) being the most frequently used predictor for hypoxia. However, deep learning and conventional ML methods are not directly comparable, as they were applied to different datasets and performance metrics were inconsistently reported. Additionally, comparability between studies was hindered by the wide variability in approaches, including the differing settings, which introduced various influences on blood oxygen saturation.

Moreover, aheterogeneous medical population is beneficial for developing a broadly applicable predictive model for hypoxia in CBRNE situations, as it increases the likelihood of achieving generalized results.

Additionally, we believe that the assumption that the model may rely on correlations and patterns among features to build its representations—potentially diverging from the established medical gold standards on which Early Warning Scores (EWS) are based—was not sufficiently developed to be considered a fundamental hypothesis.
In medicine, a "gold standard" refers to the most trusted and conventional method for diagnosing diseases, assessing treatment effectiveness, or verifying the accuracy of tests and measurements. This benchmark serves as the reference standard for comparing alternative approaches.

In this study, we used datasets from MIMIC-III and IV [29–31], employing Gradient Boosting Models [32–34] (XGBoost, LightGBM, CatBoost) and sequential models [35–37] (LSTM, GRU) to predict hypoxemia severity scores. These scores, which actually are the labels we aim to predict, were based on the newly designed NEWS2+ system adapted for pre-clinical scenarios, like CBRNE events. This adaptation, defined by medical expert annotations in the VIMY research group, includes hypoxemia severity for three population groups (adults with COPD, adults without COPD, and pediatric patients without COPD) and a modified EWS for two population groups (adult and pediatric patients). Our comprehensive preprocessing pipeline addressed missing data and class imbalances, ensuring robust and reliable model training.

This paper is structured as follows: Section 2 details the materials and methods, including data preprocessing and model development. Section 3 presents the results of our experiments. In Section 4, we discuss the implications of our findings. Finally, Section 5 concludes the study and outlines future research directions.

## 2. Materials and Methods

### 2.1. Data Sources

We used the MIMIC-III and MIMIC-IV databases [29–31], which are extensive, de-identified health records of ICU patients. MIMIC-IV is an excellent choice for beginners or those seeking to leverage well-established concepts, as it offers a robust and widely used dataset. However, while the HiRID database provides data at 2-minute intervals, MIMIC remains the most comprehensive dataset, offering greater breadth and depth for a variety of clinical research applications.

Given that, to our knowledge, no public physiological dataset has been collected specifically in a CBRNE context, we used ICU data for this proof of concept. Critical care patients frequently exhibit severe distress patterns akin to those expected in CBRNE scenarios, making this data a suitable proxy.
Furthermore, we have a very heterogeneous population, which can actually be beneficial, as our goal is to predict hypoxia in CBRNE situations in a non-specific manner. Indeed, rather than targeting a single scenario, we aim to develop a broadly applicable predictive model. Thus, a diverse population increases the likelihood of achieving broadly applicable results, enhancing the overall scope of our predictive model.
Finally, the MIMIC-III dataset includes around 58,000 hospital admissions for over 40,000 unique patients, while MIMIC-IV expands this to over 380,000 admissions for more than 210,000 patients from 2008 to 2019. Both databases offer detailed information, including continuous vital sign monitoring (recorded regularly or very frequently), laboratory results, and demographic details, making them highly applicable to our hypoxemia prediction study.

### 2.2. Inclusion Criteria and Data Representativeness

Patients were included if they had recorded episodes of hypoxemia or were diagnosed with conditions very often associated with potential low blood oxygenation levels, which made them more likely to experience hypoxemic episodes (see appendix, A6). A total of 51,368 admissions from 42,599 unique patients were selected based on the International Classification of Diseases (ICD) codes related to hypoxemia and respiratory distress. Figure 3 illustrates the simplified inclusion process.

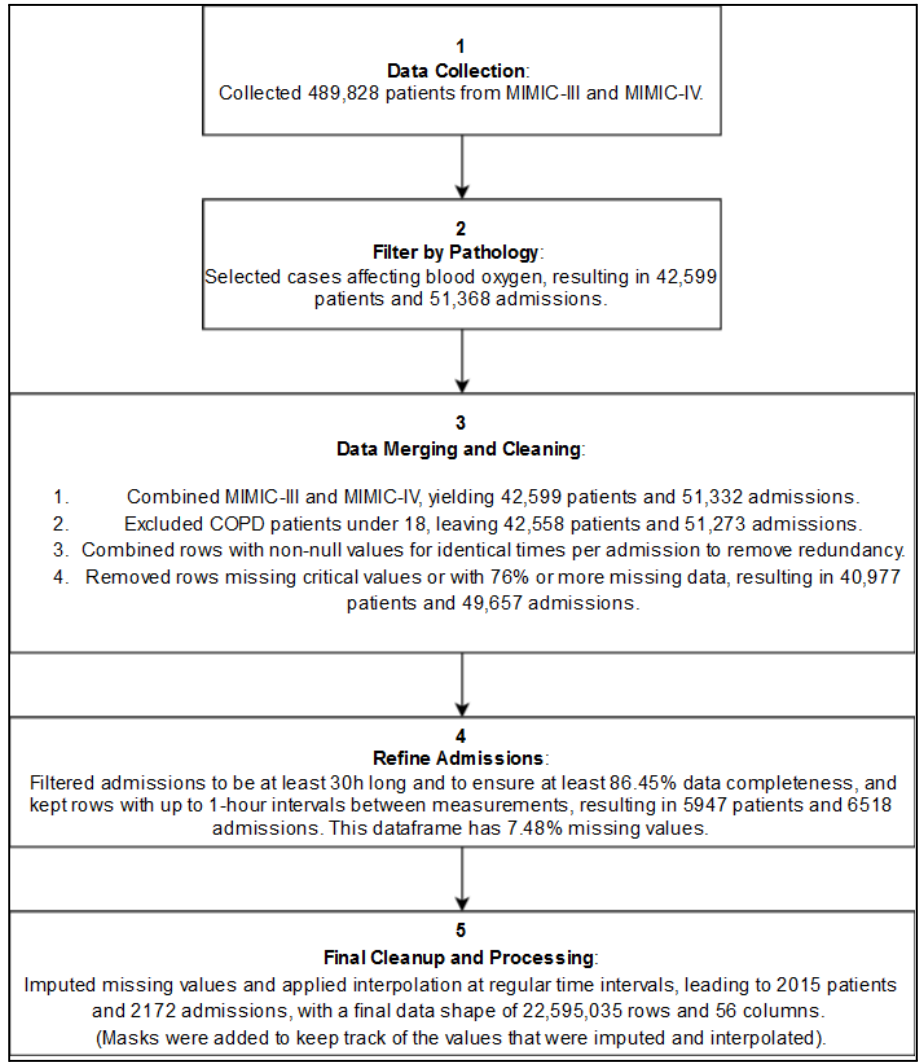

**Figure 3. Simplified Inclusion diagram showing the main preprocessing steps before the imputations and interpolations, and after.** See complete one in the appendix.

## 2.3. Data Preprocessing

### 2.3.1. Labeling and feature engineering

The VIMY project research team has initiated a new research wing focused on triage and early-warning systems to support the development of the NEWS2+ early warning system. NEWS2+ is an extension of the original NEWS2, being revisited in light of recent advancements in the scientific literature, and relevant to pre-clinical contexts. Specifically, the medical experts on the team have focused on adapting the original NEWS2 parameters for application in acute pre-hospital environments. We have developed an alternative method to "inform" the model about what constitutes a trigger value for the physiological variables in question, incorporating an alarm mechanism. Additionally, temporal information is provided for each physiological variable of interest, detailing how long each alarm persisted. Medical experts within the VIMY team have contributed significantly to this initiative, and the adapted chart will be extended for use in disaster scenarios, including CBRNE events.

***Use of SpO$_2$ for categorization.*** To categorize hypoxemia severity, we introduced a severity labeling column based on SpO$_2$ values, classifying patients as severe, moderate, mild, or normal. SpO$_2$ was chosen as the central parameter due to its prevalence and accessibility through oximeters in pre-hospital and acute care scenarios. The system assigns scores ranging from 0 (normal) to 3 (severe) based on SpO$_2$ levels, with adjustments for patient age and the presence or absence (Figure 4) of chronic obstructive pulmonary disease (COPD).

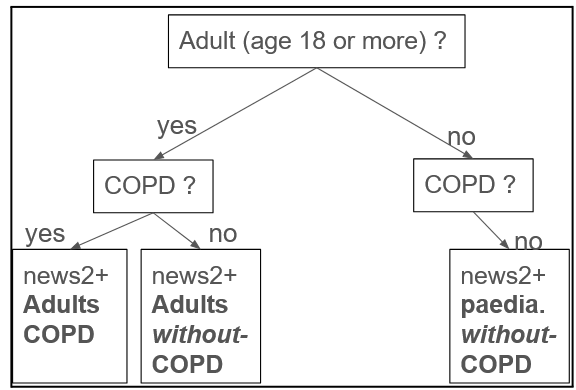

**Figure 4 Labellisation process overview**. Designed on a decision-tree like system. In our process, we have no children (patients < 18 years) with COPD, only without. For each of the three categories, we used different threshold values for SpO2 (%) to determine the severity scores.

In certain situations, gold-standard thresholds need to be adapted based on advancements in the literature. For instance, Dempsey et al. [38] described hypoxemia levels in healthy individuals as follows: mild (93-95%), moderate (88-93%), and severe (<88%). Bourassa et al [8] proposed hypoxemia thresholds of <90% for the general population and <88% for COPD patients in room air conditions [8]. They also set hypoxemia thresholds for supplemental oxygen therapy at >92% for the general population and >94% for those with COPD.
Additionally, Johannigman et al. [39] reported notable changes in SpO$_2$ during aeromedical evacuations, where 90% of military personnel experienced at least one desaturation event with SpO$_2$ <90%, and over half dropped below 85%

From the medical conditions manually selected in the MIMIC database, COPD cases were identified based on the presence of the following conditions: *Chronic Obstructive Respiratory Disease, COPD arising in the perinatal period*, *Chronic obstructive asthma with status asthmaticus*, and *Other chronic bronchitis, which could have an impact on the SpO2 level in basal state*.

The Table 1 displays the adapted values for different hypoxemia levels in adults with and without COPD, as well as in the pediatric population without COPD (pediatric population with COPD is not considered in this study). By providing tailored SpO$_2$ thresholds for varying patient needs, the NEWS2+ scoring matrices support comprehensive hypoxemia risk assessment and facilitate targeted clinical responses across diverse patient populations.

| Population | Severe | Moderate | Mild | Normal |
|---|---|---|---|---|
| Adults without COPD | ≤ 91% | 92-93% | 94-95% | 96-100% |
| Adults with COPD | ≤ 82% | 83-84% | 85-87% | 88-92% |
| Pediatric (without COPD) | ≤ 84% | 85-88% | 89-95% | 96-100% |

**Tabel 1 : NEWS2+ Scoring Matrices for Hypoxemia Severity by Patient Type.**
The NEWS2+ scoring matrices provide a structured approach to determining hypoxemia severity scores across distinct patient groups, specifically tailored for adults without COPD, adults with COPD, and pediatric populations. These scores serve as the labels we aim to predict within our study and are grounded in established research sources [4,8–13,17,20,40] , with further adaptation by physicians to suit a pre-clinical setting. The matrices provide a patient-type-specific framework for interpreting $SpO_2$ thresholds, which are essential for assessing respiratory health. $SpO_2$, or oxygen saturation, represents the percentage of oxygenated hemoglobin in the blood, a vital indicator of respiratory function. Each matrix within NEWS2+ is optimized for different oxygenation needs: adults without COPD have specific $SpO_2$ targets aligned with general adult respiratory requirements; adults with COPD have adjusted targets reflecting the lower oxygenation needs typical for COPD patients; and pediatric populations have age-specific thresholds to accommodate children's unique respiratory profiles. Each matrix categorizes $SpO_2$ levels into severity classifications to guide clinical assessment and intervention. Severe hypoxemia indicates critically low $SpO_2$, requiring immediate medical intervention. Moderate hypoxemia reflects moderate oxygen depletion, necessitating close monitoring and potential intervention. Mild hypoxemia signifies slightly low oxygen levels, usually manageable without intensive treatment. Normal levels denote stable $SpO_2$ and healthy respiratory function, not requiring intervention. .

***Feature Engineering*** The Table 2 introduces NEWS2+, an expanded version of the National Early Warning Score 2 (NEWS2), designed to provide a comprehensive assessment of patient status. Developed with guidance of the VIMY team's medical experts, NEWS2+ evaluates vital signs using specific thresholds and severity levels derived from established Gold Standards. The matrices presented outline the NEWS2+ scoring system applied to six primary physiological variables—respiratory rate, oxygen saturation ($SpO_2$), heart rate, systolic blood pressure, diastolic blood pressure, and temperature. These scores function as indicators of abnormal vital signs, supporting enhanced data interpretation and predictive modeling.
Each physiological variable is assigned a score reflecting its deviation from age-adjusted reference values, offering critical insights for hypoxemia prediction. These NEWS2+-derived feature scores, occasionally referred to as "TAG" for each physiological variable, may be used interchangeably in this work.
The Table 2 below presents the NEWS2+ matrices used to generate these score features (TAGs), demonstrating the process of assigning individual scores to each physiological variable

| Age Group | Severe Low (Level 3) | Moderate Low (Level 2) | Mild Low (Level 1) | Normal Baseline (Level 0) | Mild High (Level 1) | Moderate High (Level 2) | Severe High (Level 3) |
|---|---|---|---|---|---|---|---|
| **Adults** | | | | | | | |
| Respiratory Rate (BPM) | 0-5 | 6-7 | 8-9 | 10-20 | 21-25 | 26-29 | 30-400 |
| Saturation (SpO₂, %) | 0-84 | 85-88 | 89-95 | 96-100 | Nil: Ambient Air | Nil: Ambient Air | Nil: Ambient Air |
| Heart Rate (BPM) | 0-40 | 41-50 | 51-59 | 60-90 | 91-110 | 111-130 | 131-400 |
| Systolic BP (mmHg) | 0-90 | 91-100 | 101-114 | 115-119 | 120-129 | 130-144 | 145-300 |
| Diastolic BP (mmHg) | 0-49 | 50-59 | 60-64 | 65-79 | 80-85 | 86-90 | 91-300 |
| Temperature (°C) | 0-35.0 | 35.1-35.9 | 36.0-36.6 | 36.7-37.7 | 37.8-38.8 | 38.9-39.8 | 39.9-60 |
| **0-11 Months** | | | | | | | |
| Respiratory Rate (BPM) | ≤19 | 20-24 | 25-29 | 30-49 | 50-59 | 60-69 | ≥70 |
| Saturation (SpO₂, %) | ≤91 | 92 | 93 | 94-100 | Nil: Ambient Air | Nil: Ambient Air | Nil: Ambient Air |
| Heart Rate (BPM) | ≤99 | 100-104 | 105-109 | 110-159 | 160-164 | 165-169 | ≥170 |
| Systolic BP (mmHg) | ≤59 | 60-64 | 65-69 | 70-99 | 100-104 | 105-109 | ≥110 |
| Diastolic BP (mmHg) | ≤20 | 19-26 | 27-37 | 37-56 | 57-69 | 70-89 | ≥90 |
| Temperature (°C) | ≤34.9 | 35.3-35.4 | 35.5-35.9 | 36-37.0 | 37.1-37.4 | 37.5-37.9 | ≥38 |
| **12-23 Months** | | | | | | | |
| Respiratory Rate (BPM) | ≤19 | 20-22 | 23-24 | 25-39 | 40-49 | 50-59 | ≥60 |
| Saturation (SpO₂, %) | ≤91 | 92 | 93 | 92-100 | Nil: Ambient Air | Nil: Ambient Air | Nil: Ambient Air |
| Heart Rate (BPM) | ≤79 | 80-89 | 90-99 | 100-149 | 150-154 | 155-159 | ≥160 |
| Systolic BP (mmHg) | ≤59 | 60-64 | 65-69 | 70-99 | 100-104 | 105-109 | ≥110 |
| Diastolic BP (mmHg) | ≤20 | 19-35 | 36-40 | 41-62 | 63-70 | 71-89 | ≥90 |
| Temperature (°C) | ≤34.9 | 35.3-35.4 | 35.5-35.9 | 36-37.0 | 37.1-37.4 | 37.5-37.9 | ≥38 |
| **2-4 Years** | | | | | | | |
| Respiratory Rate (BPM) | ≤14 | 15-18 | 17-19 | 20-34 | 35-39 | 40-49 | ≥50 |
| Saturation (SpO₂, %) | ≤91 | 92 | 93 | ≥94 | Nil: Ambient Air | Nil: Ambient Air | Nil: Ambient Air |
| Heart Rate (BPM) | ≤69 | 70-79 | 80-89 | 90-139 | 140-144 | 145-149 | ≥150 |
| Systolic BP (mmHg) | ≤69 | 70-79 | 80-89 | 90-119 | 120-129 | 130-144 | ≥150 |
| Diastolic BP (mmHg) | ≤18 | 19-34 | 35-43 | 44-67 | 68-75 | 76-89 | ≥90 |
| Temperature (°C) | ≤34.9 | 35.3-35.4 | 35.5-35.9 | 36-37.0 | 37.1-37.4 | 37.5-37.9 | ≥38 |
| **5-11 Years** | | | | | | | |
| Respiratory Rate (BPM) | ≤14 | 15-16 | 17-19 | 20-29 | 30-34 | 35-39 | ≥40 |
| Saturation (SpO₂, %) | ≤91 | 92 | 93 | ≥94 | Nil: Ambient Air | Nil: Ambient Air | Nil: Ambient Air |
| Heart Rate (BPM) | ≤59 | 60-69 | 70-79 | 80-129 | 130-134 | 135-139 | ≥140 |
| Systolic BP (mmHg) | ≤79 | 80-84 | 85-89 | 90-109 | 110-119 | 120-129 | ≥130 |
| Diastolic BP (mmHg) | ≤41 | 42-47 | 48-52 | 53-79 | 80-85 | 86-89 | ≥90 |
| Temperature (°C) | ≤34.9 | 35.3-35.4 | 35.5-35.9 | 36-37.0 | 37.1-37.4 | 37.5-37.9 | ≥38 |
| **12-17 Years** | | | | | | | |
| Respiratory Rate (BPM) | ≤9 | 10-12 | 13-14 | 15-24 | 25-29 | 30-34 | ≥35 |
| Saturation (SpO₂, %) | ≤91 | 92 | 93 | ≥94 | Nil: Ambient Air | Nil: Ambient Air | Nil: Ambient Air |
| Heart Rate (BPM) | ≤49 | 50-59 | 60-69 | 70-109 | 110-119 | 120-129 | ≥130 |
| Systolic BP (mmHg) | ≤89 | 90-94 | 95-99 | 100-119 | 120-134 | 135-139 | ≥140 |
| Diastolic BP (mmHg) | ≤41 | 42-47 | 48-52 | 63-80 | 81-85 | 86-89 | ≥90 |
| Temperature (°C) | ≤34.9 | 35.3-35.4 | 35.5-35.9 | 36-37.0 | 37.1-37.4 | 37.5-37.9 | ≥38 |

**Table 2 NEWS2+ Scoring Matrix: Feature Scores (TAGs) for Physiological Variables in Adults and Pediatric Populations.** The NEWS2+ scoring system is applied to derive feature scores (TAGs) for six primary physiological variables across both adult and pediatric populations. For adults, these feature scores capture key indicators of physiological health, while the pediatric version of NEWS2+ is tailored to accommodate developmental differences across specific age groups. The pediatric matrix is divided into five distinct age groups to ensure age-appropriate scoring: infants under 1 year (0–11 months), toddlers under 2 years (12–23 months), children aged 2 to 4 years, those aged 5 to 11 years, and adolescents aged 12 to 17 years. These age groups are presented sequentially, from top to bottom, in the order listed. This structured approach ensures that the scoring accurately reflects physiological expectations and variations across developmental stages, supporting precise assessment and appropriate clinical responses.

### 2.3.2. Handling Missing Data and Masks

A comprehensive pipeline addressed missing data through imputation and interpolation methods. Missing data accounted for 7.48% of the dataset. We employed a multivariate imputation strategy: Multiple Imputation by Chained Equations (MICE) [41–43] using Histogram-based Gradient Boosting.

Synthetic data entries were flagged using masks to ensure transparency, in line with recommendations from previous research [42–45], which are critical for trust in AI-driven decision-making [46]. This approach helps inform the model about which values were imputed, improving interpretability. Specifically, each feature column had an associated mask column, where a value of 0 indicated no synthetic data and a value of 1 indicated the presence of synthetic data in that row.

To address irregular time intervals between measurements, we used interpolation to generate regular time series data. Minute-level interpolations were performed using linear interpolation, chosen after comparing it with polynomial and cubic spline methods. Linear interpolation produced the most reliable results for vital signs data, minimizing the risk of implausible physiological values (Table 3).

| Variable | Polynomial (order 3 and 5) and Cubic Spline Interpolations | | Linear Interpolations | |
|---|---|---|---|---|
| | Minimal | Maximal | Minimal | Maximal |
| Respiratory rate (breaths/min) | -73640110000 | 2.51E+14 | 0 | 121 |
| Spo2 (%) | -3782930000 | 1.81E+14 | 0 | 100 |
| Heart rate (bpm) | -8.62E+13 | 2.52E+10 | 0 | 268 |
| Blood pressure systolic (mmhg) | -1.22E+11 | 3.82E+10 | 0 | 261 |
| Blood pressure diastolic (mmhg) | -73027180000 | 5.74E+08 | 0 | 228 |
| MAP(mmhg) | -83183000000 | 1.07E+10 | 0 | 238 |

**Table 3. Comparison of interpolation methods showing physiological variables' maximum range values after interpolation**. We observed a typical distribution of maximum values resulting from polynomial interpolations of orders 3 and 5, noting the occurrence of negative values—anomalous in this context—as well as an implausible order of magnitude. Similar issues were found with cubic spline interpolations. In contrast, linear interpolation displayed a typical distribution in which all values remained positive, with an order of magnitude that appropriately reflected the extreme physiological conditions possible in ICU settings. Therefore, we ultimately selected linear interpolation for our analyses.

We added a mask for the charttimes —the timestamps (at the minute) corresponding to when each set of measurements was recorded— indicating which rows were interpolated, to inform the models about synthetic data (same logic as for the imputed values before). This approach aligns with the intuition that linear interpolation can effectively represent gradual changes in a patient's condition between successive measurements.

### 2.3.4. Sliding window

To determine an appropriate time window for prediction, we analyzed the average duration of each hypoxemia severity level after interpolation. The mean and median durations for severity scores 0 to 3 were calculated, with severity score 3 having a mean duration of 114.8 minutes and a median of 60.0 minutes. Drawing from these observations, practical considerations, discussions with our team's medical expert, and findings from previous studies [28,47,48], we established a 5-minute prediction window as the standard for our models. This timeframe is long enough to allow doctors sufficient time to intervene, while remaining short enough to capture near-term risks that require immediate attention. Additionally, it minimizes the risk of inaccurately predicting events that are too far in the future. This balance ensures timely and actionable medical interventions.

### 2.3.6. Data Cleaning and Transformation

- **Duplicated Rows and Admissions Removal:** To address potential redundant rows introduced during preprocessing, we merged rows with non-null values recorded at the same charttime (i.e., the same timestamp) within each admission. For eventual duplicate admissions across the combined MIMIC III and IV datasets (which may have different admission IDs), only the most recent occurrence of each redundant row (considering all physiological and demographic features) was retained, while earlier rows were removed to maintain data consistency. This approach ensures that, for each potentially redundant admission, only a single, complete version remains in the final dataset, effectively eliminating duplicate admissions on a row-by-row basis.
- **Outlier Handling:** Implausible physiological measurements were replaced with NaN and subsequently imputed to address potential human errors in the electronic health records. Specifically, for six key vital signs variables, we excluded values outside defined ranges: respiratory rate, heart rate, and both systolic and diastolic blood pressure above 300 or below 0; $SpO_2$ above 100 or below 0; and temperature readings above 60 or below 0. To ensure consistency, this step was applied both before and after imputation and interpolation.
- **Data Rounding:** Physiological values were rounded in order to standardize precision, which supports model generalization and may reduce the likelihood of overfitting, especially as future work integrates data from additional hospitals.
- **Missing values**: Imputation by Chained Equations (MICE) using Histogram-based Gradient Boosting were done to complete the 7.48% missing values from our preprocessed dataframe
- **Time Alignment:** Interpolated data at minute-level intervals to standardize time steps across admissions (synthetic data).

### 2.3.7. Data Splitting

To ensure robust comparisons and avoid data leakage, patient data was split into training, validation, and testing sets, ensuring no patient appeared in more than one set (patient wise not admission wise). Specifically, 75% of patients were allocated to the training set (2,015 patients), while the validation and test sets each comprised 12.5% of the patients (212 patients each).

## 2.4. Exploratory Data Analysis (EDA)

We conducted EDA to understand data distributions and correlations:

- **Demographics:** Table 4 illustrates the patients' age, gender, race, and ethnicity distributions.

    Concerning the demographics, for the races and ethnicities, we referred to the United States Census Bureau[49].

    The age stratifications shown below were a constitution work from various sources [50,51] and were used to do the NEWS2+ derived features scores (TAGS) for the six key physiological parameters.

    | Category | Subcategory (years) | Count | Percentage (%) |
    |---|---|---|---|
    | Age Stratification | <1 year | 0 | 0 |
    | | 1 to <2 | 0 | 0 |
    | | 2 to 4 | 0 | 0 |
    | | 5 to 11 | 0 | 0 |
    | | 12 to 17 | 4 | 0.061369 |
    | | 18 to 45 | 836 | 12.826202 |
    | | 46 to 65 | 2369 | 36.345505 |
    | | 66 to 85 | 2693 | 41.316355 |
    | | 86+ | 616 | 9.450752 |
    | Gender | Male | 3511 | 53.866217 |
    | | Female | 3007 | 46.133783 |
    | Race/Ethnicity | White | 4652 | 71.371586 |
    | | Undefined | 694 | 10.647438 |
    | | Black / African American | 686 | 10.524701 |
    | | Hispanic / Latino | 259 | 3.973612 |
    | | Asian | 199 | 3.053084 |
    | | American Indian / Alaska Native | 13 | 0.199448 |
    | | Native Hawaiian / other Pacific Islander | 9 | 0.138079 |
    | | Multiracial | 6 | 0.092053 |

    **Table 4. Demographics of the study population after preprocessing.** The top panel illustrates age stratification across admissions, revealing no pediatric patients, four teenage admissions, and predominantly adult admissions. Most adults are between the ages of 45-65 or 66-84, with a slightly higher proportion of elderly individuals (66+). Age groups are categorized as infant (0-17 years), adult (18-65 years), and elderly (65+ years). The middle panel shows gender distribution, with a moderate overrepresentation of males compared to females. The bottom panel presents the distribution of race and ethnicity, highlighting an overrepresentation of white individuals. Here, "Hispanic/Latino" is categorized as an ethnicity, while other categories are classified by race.

- **Label Distribution:** We observed an unbalanced dataset, where higher severity scores occur less frequently (Table 5). The labels, defined as 0 (normal), 1 (mild), 2 (moderate), and 3 (severe), each represent a specific hypoxemia severity score. They are based on the NEWS2+ scoring system in respect to the Spo2, age and type of disease (see feature engineering, part 2.3.1).

| Label (Hypoxemia Severity Score) | Before Interpolation | | After Interpolation | |
|---|---|---|---|---|
| | Count | Percentage (%) | Count | Percentage (%) |
| 0 | 729657 | 72.2369 | 17450087 | 76.86217 |
| 1 | 158419 | 15.68367 | 3293482 | 14.50676 |
| 2 | 73338 | 7.260548 | 1244804 | 5.482972 |
| 3 | 48675 | 4.818882 | 714716 | 3.1481 |

**Table 5. Label distribution of hypoxemia severity scores before and after interpolation.** Left columns : before; Right columns : after. The count numbers increase due to the minute-by-minute interpolations adding more rows. However, the label distribution, in terms of percentage, remains very similar and retains its characteristics, making it suitable for later training. The imputations (done before interpolations) and the interpolations did not change the label distribution.

- **Correlation Analysis:** We generated correlation matrices to assess relationships between physiological variables (Figure 5).

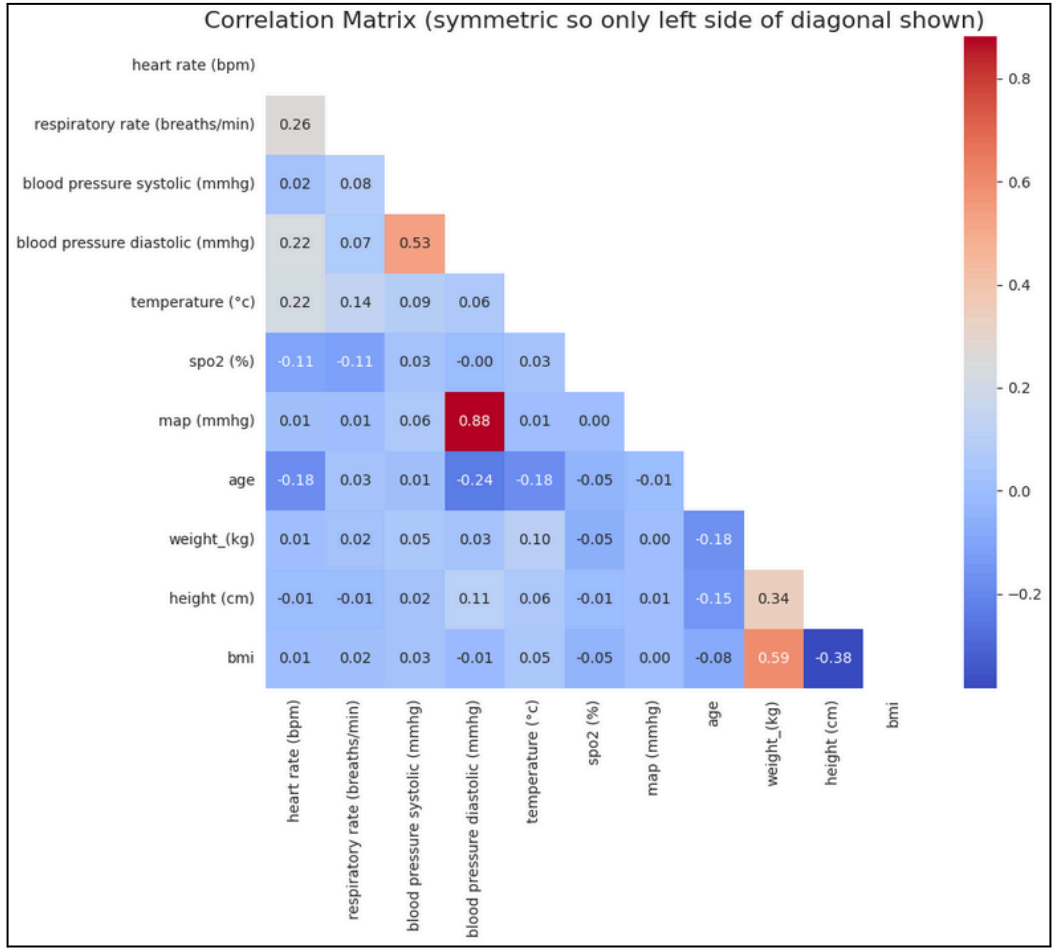

**Figure 5. Correlation matrix of main physiological variables before imputations and interpolations.** There is a strong positive correlation between mean arterial pressure (MAP) and diastolic blood pressure (BP), which is expected since MAP is derived from both diastolic and systolic BP (mathematical relation). Similarly, a strong positive correlation between systolic and diastolic BP aligns with physiological norms. As anticipated, weight and BMI are also strongly correlated, given that BMI is calculated from weight. Conversely, a negative correlation between BMI and height is observed, as BMI decreases with increasing height. Additionally, within a certain age range, a negative correlation between age and diastolic BP can be expected. These correlations underscore the physiological relationships between these variables. Both BMI and MAP are physiologically significant and may provide valuable insights for predicting the risk or severity of hypoxemia.

## 2.5. Principal Component Analysis (PCA)

PCA was performed after imputation and interpolation to assess the variance explained by each principal component and feature contributions. Due to PCA's sensitivity to data scaling, we applied standardization beforehand. The cumulative variance from the first few components indicated that a reduced feature set could capture most data variability (Figure 6). We focused on physiological variables and easily obtainable features (e.g., gender, height, weight) relevant to CBRNE disaster scenarios, excluding other features for this analysis. While PCA is a linear technique and medical data can show non-linear properties, the results provide useful insights into our physiological features, though they should be interpreted cautiously.

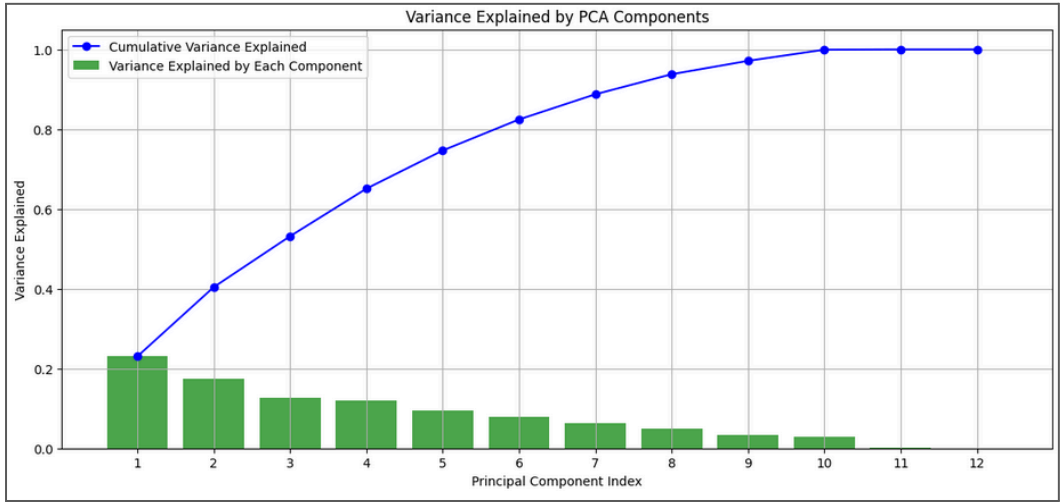

**Figure 6. PCA variance explained and feature contributions.** In green, the percentage of variance accounted for by each principal component is illustrated, and in blue, the cumulative variance of each PC added to those before.

## 2.6. Model Development

### 2.6.1. Model Selection and Rationale

We selected five models for comparison: Gradient Boosting Models (XGBoost [34], LightGBM [32], CatBoost [33]) and sequential models (LSTM [37] and GRU [36]). The choice was motivated by recent studies demonstrating the effectiveness of both GBMs and sequential models in handling medical time series data [52–56].

The Gradient Boosting Models (GBMs) are effective for tabular data and can handle complex, non-linear relationships. The Sequential Models are designed to capture temporal dependencies in sequential data.

### 2.6.2. Gradient Boosting Models (GBMs)

We implemented XGBoost, LightGBM, and CatBoost classifiers. These models are ensemble methods that build upon decision trees, focusing on correcting errors of previous models in the sequence. They offer advantages such as handling missing values and providing feature importance (Figure 10). Note that the shift lag process for the labels was applied for both GBMs and sequential models (see the Figure 8 for the shift lag and 2.6.4 part for further explanations).

The GBMs each offer unique features suited to specific modeling challenges. For instance, XGBoost applies early stopping and class weights to address class imbalance, CatBoost handles categorical variables effectively and reduces prediction shifts, and LightGBM is optimized for speed with Gradient-based One-Side Sampling.

| Model Classifier (Multi-class) | Setup (Default Hyperparameters) | Tuned Hyperparameters (HyperOpt library) |
|---|---|---|
| XGBoost | - Objective: 'multi:softmax' function<br>- Number of Classes: 4<br>- Random Seed: 42<br>- Evaluation Metric: 'mlogloss'<br>- Number of Boosting Rounds: 300<br>- Early Stopping Rounds: 5<br>- Device: GPU | - max_depth: 4 to 8<br>- eta: 0.005 to 0.3<br>- subsample: 0.5 to 1<br>- gamma: 0 to 22<br>- min_child_weight: 0 to 15 |
| CatBoost | - Objective: 'MultiClass' function<br>- Random Seed: 42<br>- Iterations: 400<br>- Evaluation Metric: 'MultiClass'<br>- Early Stopping Rounds: 5<br>- Device: GPU | - depth: 4 to 10<br>- learning_rate: 0.005 to 0.3<br>- l2_leaf_reg: 1 to 10<br>- min_data_in_leaf: 1 to 15 |
| LightGBM | - Objective: 'multiclass' function<br>- Number of Classes: 4<br>- Random Seed: 42<br>- Number of Boosting Rounds: 300<br>- Device: GPU | - max_depth: 4 to 8<br>- learning_rate: 0.005 to 0.3<br>- bagging_fraction: 0.5 to 1<br>- min_split_gain: 0 to 22<br>- min_child_weight: 0 to 15 |

**Table 6: Default and tuned hyperparameters of the three gradient-boosting machine models (GBMs).** The hyperparameters were optimized using the HyperOpt library and subsequently added to the existing default parameters.This table presents the search ranges of the hyperparameters used to identify the optimal combination for each model (see Table 9 for the best results).

### 2.6.3. Sequential Models

We implemented Masked LSTM and Masked GRU models to process sequential data, addressing variable sequence lengths through padding and the masking process.

More specifically, among the 2,169 unique admissions in the three datasets, sequence lengths—representing the duration of each admission—ranged from 952 to 254,643 rows, with a median of 5,941 rows and an average of 10,417 rows (each row corresponding to one minute per admission). Thus, a significant difference in terms of the length.
To ensure computational efficiency of the sequential models, we standardized the sequence length to 1,024 rows—about 18 hours per admission. Longer sequences were split into 1,024-row segments, while shorter ones were padded with a constant value of 1000, which didn't overlap with any actual feature values. This preprocessing generated 23,099 admissions suitable for sequential models.

The Masked LSTM and GRU models were designed to ignore padded values (set to 1000) using an internal mask within the Dataloader. This mask excludes padded values in sequential input data, ensuring they do not influence model training. Notably, this type of "masking" differs from the masks used to identify imputed or interpolated (synthetic) data (see Section 2.3.2).

Variable-length sequences are effectively managed through masking and sequence packing. Sequence lengths are calculated from the mask, and sequences are sorted by length. Using

PyTorch's pack_padded_sequence, sequences are packed so that LSTM/GRU layers only process valid time steps, omitting padding from computations. After processing, sequences are unpacked and returned to their original order, enabling efficient batch processing of variable-length sequences and ensuring that padding does not interfere with the model's learning.

See Figure 7 for an overview of the preprocessing steps for the sequential models.

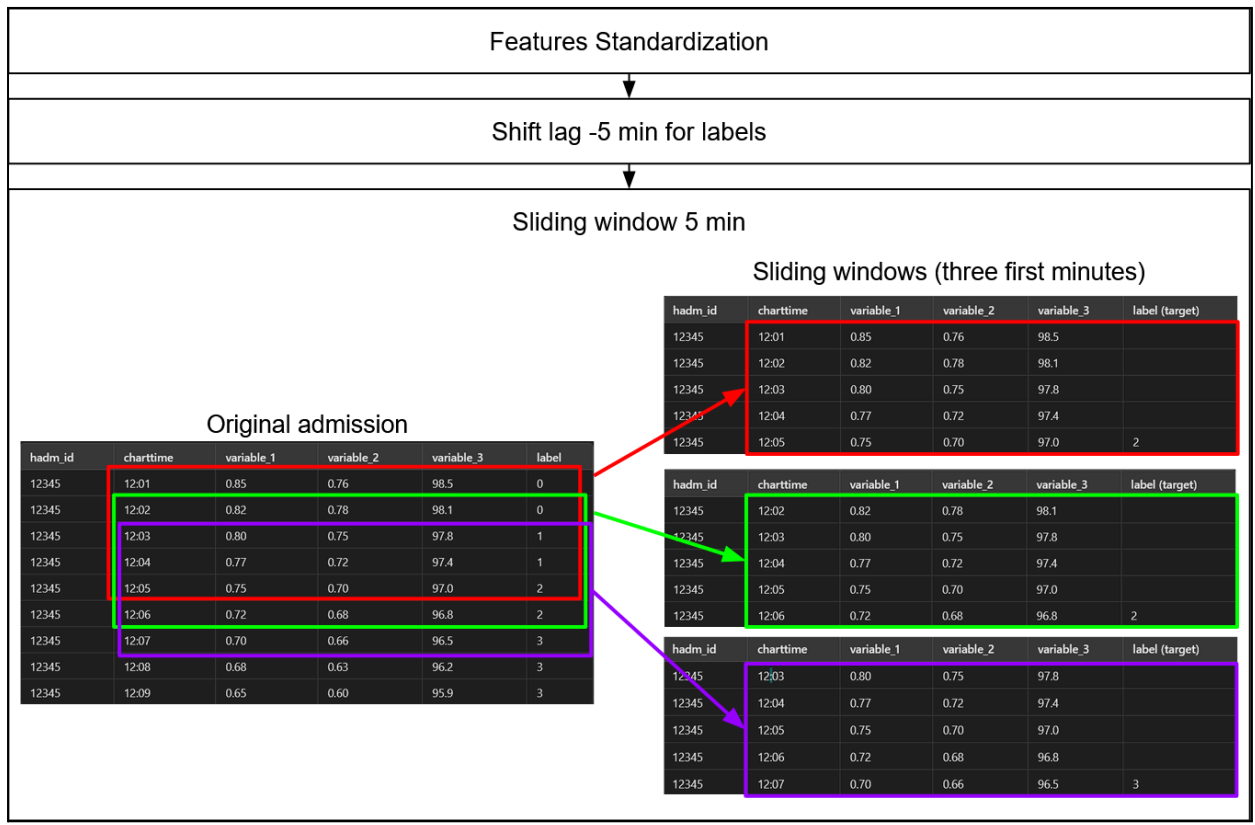

**Figure 7: Overview of Main Preprocessing Steps for Sequential Models with a 5-Minute Sliding Window Illustration for the Initial Minutes of an Admission. Step 1** : involves standardizing the data using Scikit-Learn's StandardScaler, but only for our sequential models. Models based on decision trees, such as gradient-boosted machines (GBMs), operate based on threshold values, so standardization is unnecessary for them. **Step 2** : we apply a -5-minute shift to the label values, pairing each row with the label occurring 5 minutes later. This adjustment allows each row, representing a set of physiological measurements at a specific minute of a patient's admission, to serve as the input for predicting the hypoxemia severity score 5 minutes in advance. Notably, this label-shifting process is performed for both sequential models and GBMs. **Step 3** : we create 5-minute sliding windows for our sequential models only, as GBMs do not leverage temporal sequences. Each sequential model therefore considers a 5-minute sequence of admission data at a time to predict the hypoxemia severity score label 5 minutes ahead. We then "jump" forward by one row (or one minute), creating a 4-minute overlap between consecutive windows, and repeat this process until we reach the end of the data for each patient admission.

Additional characteristics of the models are presented in Table 7 below.

| Model Classifier | Architecture and Hyperparameters | Training Configuration |
|---|---|---|
| LSTM | - Input Processing:<br>  - Input features are masked: x = x * mask<br>  - Variable sequence lengths are handled via packing/unpacking sequences<br>- **Recurrent Layers**:<br>  - **3-layer LSTM** with:<br>    - Input Size: 41 (number of features)<br>    - Hidden Size: 256 units per layer<br>    - Batch First: True<br>- Fully Connected Layers:<br>  - FC1: Linear layer with 256 input and 256 output units<br>  - Activation Function: ReLU<br>  - FC2: Linear layer with 256 input and 4 output units (number of classes)<br>- **Total Trainable Parameters**: 1,425,668 | |
| GRU | - Input Processing:<br>  - Input features are masked: x = x * mask<br>  - Variable sequence lengths are handled via packing/unpacking sequences<br>- **Recurrent Layers**:<br>  - **3-layer GRU** with:<br>    - Input Size: 41 (number of features)<br>    - Hidden Size: 256 units per layer<br>    - Batch First: True<br>- Fully Connected Layers:<br>  - FC1: Linear layer with 256 input and 256 output units<br>  - Activation Function: ReLU<br>  - FC2: Linear layer with 256 input and 4 output units (number of classes)<br>- **Total Trainable Parameters**: 1,085,956 | - Batch Size: 64<br>- Learning Rate: 0.001<br>- Number of Epochs: 15 (optimal performance achieved earlier)<br>- Weight Decay: 0.0001<br>- Training Method: 5-minute sliding window with a shift lag of -5 minutes<br>- Device: GPU (T4 or L4 GPU used) |

**Table 7: Architecture and hyperparameters of the sequential models**. The "masked" layers use an internal Dataloader mask to exclude padded values in sequential input data, ensuring they are not considered during training. Variable-length sequences are handled through packing and unpacking: a mask ignores padding, and *pack_padded_sequence* enables LSTM/GRU layers to process only valid steps. After processing, sequences are unpacked to their original order, preventing padding from affecting learning. Due to long training times, hyperparameters were not extensively optimized; instead, a sufficiently complex architecture was implemented, aimed at fitting the training data well enough to demonstrate proof of concept.

### 2.6.4. Experiments' Setup

To prevent data leakage, we ensured that each patient's data appeared in only one of the training, validation, or testing sets. Admissions were split accordingly.

- **Shift-Lag Method:** Applied to predict hypoxemia severity scores 5 minutes in advance, providing a practical window for medical intervention (Figure 8). Indeed, in order to predict hypoxemia severity scores in advance, we apply a -5-minute shift to the label values, aligning each row with the label observed 5 minutes later. This adjustment allows each row, containing physiological measurements recorded at a specific minute of a patient's admission, to be used as input for predicting the hypoxemia severity score 5 minutes ahead. Importantly, this label-shifting technique is consistently implemented for both sequential models and gradient-boosting machines (GBMs).

| Time | Y before | Y after |
|---|---|---|
| 8h15min | 316.1 | 317.3 |
| 8h16min | 317.3 | 317.6 |
| 8h17min | 317.6 | 317.5 |
| 8h18min | 317.5 | 316.4 |
| 8h19min | 316.4 | 316.9 |

**Figure 8. Shift-lag illustration for a shift of -1 (for illustrative purposes).** Observe the red arrows to see how each y label is moved one position backward (denoted as -1). The result : the label of each given row is the future severity in x-minutes. (Where label "Y" represents, in our case, the hypoxemia severity score).

- **Class Imbalance Handling:** Computed class weights inversely proportional to class frequencies.
- **Hyperparameter Tuning of the GBMs:** We used the HyperOpt library, a popular tool for hyperparameter optimization, with two objective functions—AUC-based and log loss-based. HyperOpt employs the Tree-structured Parzen Estimator (TPE), a Bayesian optimization technique that iteratively updates a probabilistic model based on prior evaluations and performance metrics. Unlike traditional methods that treat these evaluations as independent, TPE refines its model progressively to capture the relationship between hyperparameters and performance, enhancing optimization efficiency.
  - **AUC-Based**: This function focuses on maximizing the AUC score by minimizing 1 - AUC, which is especially beneficial when dealing with class imbalance.
  - **Log Loss**: This function minimizes log loss (cross-entropy loss), assessing how closely the predicted probabilities match the actual labels, and penalizing incorrect or overly confident predictions.

#### 2.6.5. Feature Importance Analysis

For the GBMs, the feature importance was assessed to understand the contribution of each feature to the models' predictions. Both TAGs derived from NEWS2+ (Table 2) and mask features indicating synthetic data were included (no masks for the CatBoost models). This analysis helps in interpreting GBMs decisions and ensuring that critical features are appropriately weighted.

## 3. Results

### 3.1. Dataset Characteristics

After preprocessing, the dataset consisted of 22,595,035 rows and over 40 feature columns, encompassing both original and engineered features. These features included physiological variables, NEWS+ derived scores (TAGs), demographic data, and mask features (the list of features is available in Table 8 below).

Each admission was interpolated to minute-level intervals, resulting in a substantial increase in data volume. The length of admissions varied, ranging from approximately 16 hours to 177 days for the GBMs. However, for the sequential models, all admission durations were standardized to approximately 16 hours by padding shorter admissions and truncating longer ones.

| Feature Name | Description | Measure |
|---|---|---|
| Gender | Biological gender of the patient. | N/A |
| Age | Age of the patient. | Years |
| Weight | Body weight of the patient. | Kilograms (kg) |
| Height | Body height of the patient. | Centimeters (cm) |
| BMI | Body Mass Index, widely used metric to assess body weight relative to height, providing a quick estimate of body fatness for most individuals. | N/A |
| Blood Pressure Systolic | Systolic blood pressure of the patient. | mmHg |
| Blood Pressure Diastolic | Diastolic blood pressure of the patient. | mmHg |
| MAP | Mean Arterial Pressure (MAP), an important cardiovascular metric representing the average pressure in a person's arteries during one cardiac cycle. | mmHg |
| Temperature | Body temperature of the patient. | Degrees Celsius (°C) |
| Heart Rate | Number of heartbeats per minute. | Beats per minute (bpm) |
| Respiratory Rate | Number of breaths per minute. | Breaths per minute |
| Oxygen Saturation ($SpO_2$) | Oxygen level in the blood. | Percentage (%) |
| Ethnicity or Race | Ethnicity or race of the patient, including Native Hawaiian / Other Pacific Islander, American Indian / Alaska Native, Black / African American, Asian, Multiracial, Hispanic / Latino (classified as an ethnicity), and White. | N/A |
| TAG_ | Derived feature scores based on NEWS2+ for six key physiological variables: TAG_Heart_Rate, TAG_Temperature, TAG_$SpO_2$, TAG_Diastolic_BP, TAG_Systolic_BP, TAG_Respiratory_Rate. | N/A |
| Mask_ | Indicates if values are synthetic (i.e., interpolated or imputed, by 1) or original (by 0) for the following variables: $SpO_2$, Systolic_BP, Diastolic_BP, Respiratory Rate, Temperature, BMI, Heart Rate, MAP, Height, Weight, and the six TAG_. | N/A |
| Label (output to be predicted, as a classification task) | Hypoxemia severity score based on the NEWS2+ scoring matrices for adults without COPD, adults with COPD, and pediatric populations (see appendix A1-A3 for scoring matrices). | N/A |

**Table 8 : Common feature columns** used to train both GBMs and sequential models. Notice that there are also masks for the six different TAGs-score-features.

## 3.2. Imputation and Interpolation Outcomes

Different imputation strategies were compared. The Histogram-based Gradient-boosting imputation yielded satisfactory results and was selected for further analysis. Imputations reduced the percentage of missing values to zero, enabling subsequent interpolation at the minute level.

Linear interpolation was chosen over polynomial and cubic spline interpolations due to its ability to produce plausible physiological values (Table 3). Polynomial and cubic spline interpolations resulted in implausible negative values and incorrect magnitudes.

## 3.3. Exploratory Data Analysis Findings

The dataset exhibited an inherent imbalance, with the distribution of severity scores remaining consistent both before and after interpolation. More severe scores were less frequent, reflecting this imbalance. In terms of label durations, severity score 2 had the shortest median duration of 29 minutes, while severity score 0 had the longest median duration, lasting 178 minutes.

## 3.4. Model Performance

### 3.4.1. Gradient Boosting Models' Results and Discussion

The GBMs demonstrated strong performance in predicting hypoxemia severity. Hyperparameter tuning yielded only slight improvements over baseline models.

**Training and Convergence:**

All models converged rapidly, with XGBoost models converging in fewer than 65 iterations. CatBoost and LightGBM showed good generalization, sometimes performing better on the validation set.

**Best Hyperparameters:**

| Model | Objective Function | Tuned Hyperparameters (post fine-tuning) |
|---|---|---|
| XGBoost | AUC | - eta: 0.2534<br>- gamma: 5.1162<br>- max_depth: 8<br>- min_child_weight: 4.6473<br>- subsample: 0.8244 |
| LightGBM | AUC | - bagging_fraction: 0.6408<br>- learning_rate: 0.2269<br>- max_depth: 5<br>- min_child_weight: 4.2377<br>- min_split_gain: 5.8370 |
| CatBoost | Logloss | - depth: 10<br>- l2_leaf_reg: 8.1064<br>- learning_rate: 0.1083<br>- min_data_in_leaf: 14.9935 |

**Table 9: Best fine-tuned models for each GBM following optimization.** This Table shows the objective functions under which each model achieved optimal performance on the metrics, along with the best hyperparameter values identified during optimization using HyperOpt-TPE, which define the final models. AUC (Area Under ROC Curve). The name of the hyperparameters are shown as they are. Refer to Table 6 for the range of hyperparameter values explored as the search space for fine-tuning the different models.

**Performance Metrics :**

The results below provide a performance benchmark for the best fine-tuned models. Notably, for all three GBMs, performance differences between each fine-tuned model and its respective baseline (default hyperparameters) were minimal, with metrics remaining closely aligned.

| Model Configuration | Accuracy | Precision Average | Sensitivity Average | Specificity Average | F1-Score Average | Macro Avg Precision | Macro Avg Sensitivity | Macro Avg F1 | Weighted Avg Precision | Weighted Avg Sensitivity | Weighted Avg F1 | MCC | AUROC Average | AUPRC Average |
|---|---|---|---|---|---|---|---|---|---|---|---|---|---|---|
| XGBoost Baseline | 0.95 | 0.85 | 0.9075 | 0.9825 | 0.87 | **0.85** | **0.91** | **0.87** | **0.96** | 0.95 | **0.96** | 0.8882 | **0.995** | **0.9325** |
| XGBoost Opti AUC | 0.95 | 0.85 | 0.9075 | 0.9825 | 0.87 | **0.85** | **0.91** | **0.87** | 0.95 | 0.95 | 0.95 | 0.8851 | **0.995** | **0.9325** |
| XGBoost Opti LogLoss | 0.95 | 0.85 | 0.9075 | 0.9825 | 0.87 | **0.85** | **0.91** | **0.87** | **0.96** | 0.95 | **0.96** | 0.888 | **0.995** | **0.9325** |
| CatBoost Baseline | **0.96** | 0.85 | **0.91** | 0.9825 | 0.87 | **0.85** | **0.91** | **0.87** | **0.96** | 0.95 | **0.96** | 0.8912 | **0.993** | **0.93** |
| CatBoost Opti AUC | **0.96** | 0.85 | 0.9075 | 0.9825 | 0.87 | **0.85** | **0.91** | **0.87** | **0.96** | **0.96** | **0.96** | 0.8894 | **0.995** | **0.9325** |
| CatBoost Opti LogLoss | 0.95 | 0.85 | 0.9075 | 0.975 | 0.87 | **0.85** | **0.91** | **0.87** | **0.96** | **0.96** | **0.96** | 0.8866 | **0.995** | **0.9325** |
| LightGBM Baseline | 0.95 | 0.85 | 0.885 | 0.9825 | 0.87 | **0.85** | **0.91** | **0.87** | **0.96** | 0.95 | **0.96** | 0.8878 | **0.995** | **0.9325** |
| LightGBM Opti AUC | 0.95 | 0.85 | 0.905 | 0.9525 | 0.87 | **0.85** | **0.91** | **0.87** | 0.95 | 0.95 | 0.95 | 0.8854 | **0.995** | **0.9325** |
| LightGBM Opti LogLoss | 0.95 | 0.85 | 0.885 | 0.9525 | 0.87 | **0.85** | **0.91** | **0.87** | **0.96** | **0.96** | **0.96** | 0.8866 | **0.995** | **0.9325** |

**Table 10: Comparison of performance metrics for the GBMs (XGBoost, CatBoost, and LightGBM models) across various configurations**. Some metrics are presented as averages, where "average" represents the mean result across all four classes for each metric, providing a streamlined overview (for detailed class-specific metrics, see the appendix part). Consistently high scores underscore the models' effectiveness in multi-class classification tasks. Optimization of objective functions (AUC and LogLoss) using the HyperOpt library yields slight performance gains, highlighting the impact of hyperparameter tuning. Optimized models are indicated with "Opti" as well as the objective function used for this goal. *Note:* Sensitivity is equivalent to Recall.

**Feature Importance Analysis:**

For XGBoost and LightGBM, both models considered a wide range of features, including physiological variables, TAGs, and mask features. The top contributing features were **heart rate, systolic blood pressure, age, weight, and respiratory rate**.

In contrast, CatBoost placed more emphasis on TAG features, with mask features being less significant. The top features in CatBoost included **$SpO_2$, TAG $SpO_2$, height, age, and heart rate** (see Figure 10).

**Confusion Matrices:**

The hypoxemia severity scores are the labels. Labels 2 and 3 were frequently misclassified, indicating challenges in predicting transitions between severity levels. Label 0 was most accurately predicted, followed by labels 3,1 and 2 (see label distribution, part 2.4). This aligns with the inherent challenge of predicting events that are both sudden and rare, such as the label "2." Importantly, this indicates that we can predict our primary complication, label 3, which corresponds to the most severe hypoxemic condition, with a reasonable degree of accuracy, supporting our proof of concept (POC).

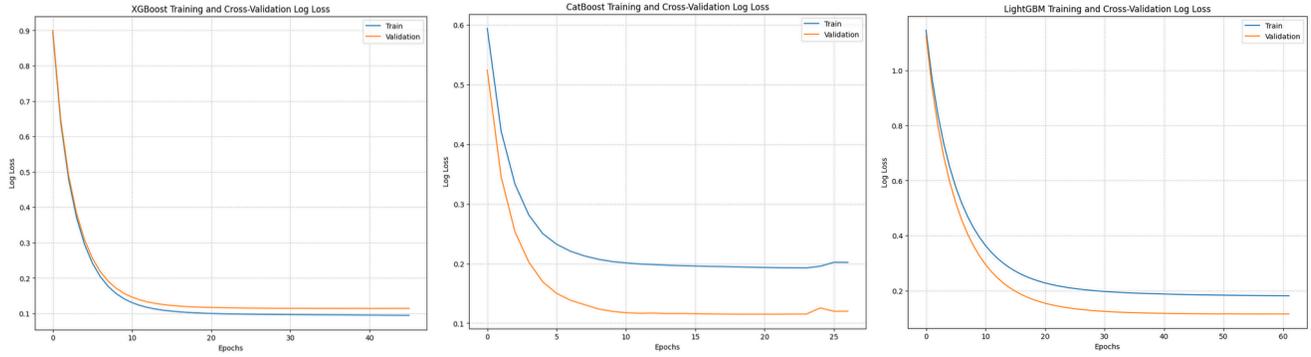

**Figure 9.** Training curves of baseline models showing loss over iterations for the 5 min in advance predictions. XGBoost (left), CatBoost (middle) and LightGBM (right). XGBoost showed slight overfitting, while CatBoost and LightGBM generally demonstrated better generalization to the validation set compared to the training set. Given the overall good performance of all models, we concluded that they all learned effectively during training.

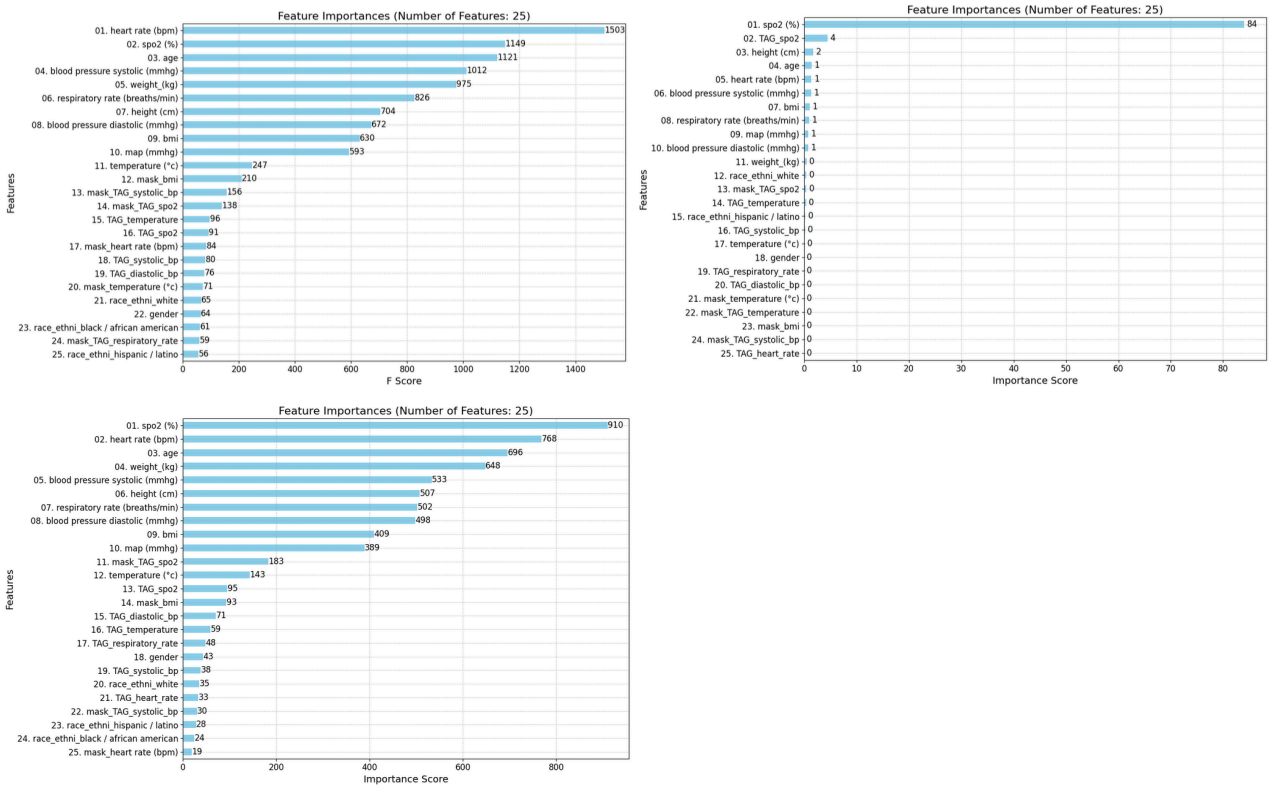

**Figure 10 : Top 25 Feature Importance derived from the GBMs for predicting hypoxemia score 5 minutes in advance:** XGBoost (top left), CatBoost (top right), and LightGBM (bottom left). The models do not utilize the features in the same way when constructing their successive prediction trees. This discrepancy highlights the differences in how each model prioritizes and processes the various features to make predictions.

### 3.4.2. Sequential Models' Results and Discussion

The LSTM and GRU models were trained using a 5-minute sliding window, with padding and masking applied to handle variable sequence lengths. By saving the models' parameters and hyperparameters at each epoch, we were able to reload the model from the optimal training epoch, ensuring the best performance was retained post-training.

**Performance Metrics:**

| Model Configuration | Accuracy | Precision Average | Sensitivity Average | Specificity Average | F1-Score Average | Macro Avg Precision | Macro Avg Sensitivity | Macro Avg F1 | Weighted Avg Precision | Weighted Avg Sensitivity | Weighted Avg F1 | MCC | AUROC Average | AUPRC Average |
|---|---|---|---|---|---|---|---|---|---|---|---|---|---|---|
| LSTM Model | 0.95 | 0.8661 | 0.9015 | 0.9821 | 0.8768 | **0.85** | **0.91** | **0.87** | **0.96** | 0.95 | **0.96** | 0.8907 | 0.995 | 0.9325 |
| GRU Model | 0.95 | **0.872** | 0.9016 | 0.9853 | 0.8799 | **0.85** | **0.91** | **0.87** | **0.96** | **0.96** | **0.96** | 0.898 | 0.995 | 0.945 |

**Table 11: Comparison of performance metrics for sequential models (LSTM and GRU) across various configurations.** Some metrics are presented as averages, where "average" denotes the mean result across all four classes, providing a concise overview (for detailed class-specific metrics, see the appendix). Consistently high scores highlight these models' effectiveness in multi-class classification tasks. Due to extensive training times, no fine-tuning was performed; instead, architectures with sufficient complexity were used as a proof of concept in this study. *Note:* Sensitivity is equivalent to Recall.

**Training Time:**

Training time was significantly longer than GBMs. Each training epoch took approximately 2.5 hours. Total training time was around 37.5 hours for each sequential model.

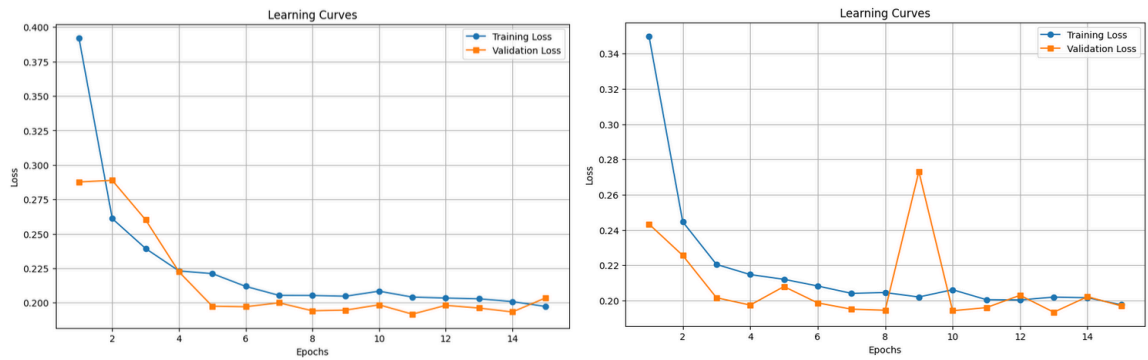

**Figure 11.** Training curves over epochs for LSTM (left) and GRU (right) models.

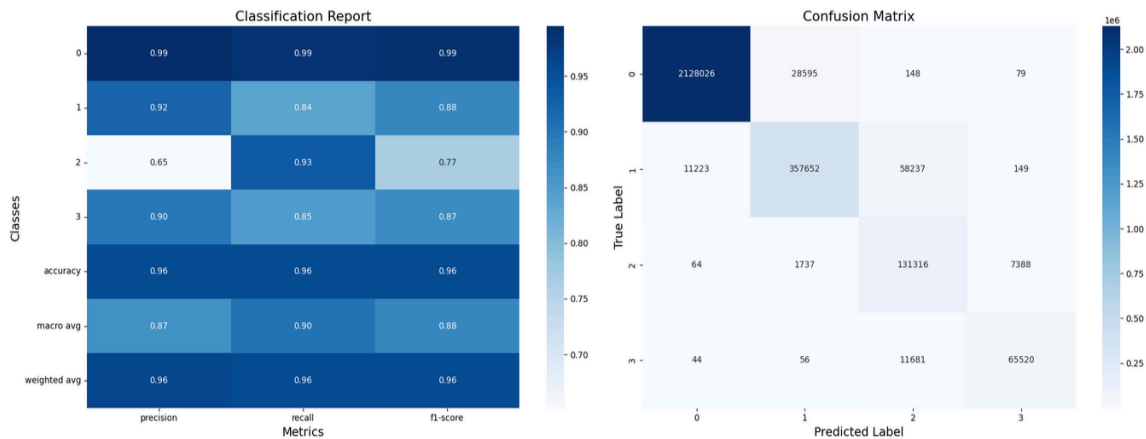

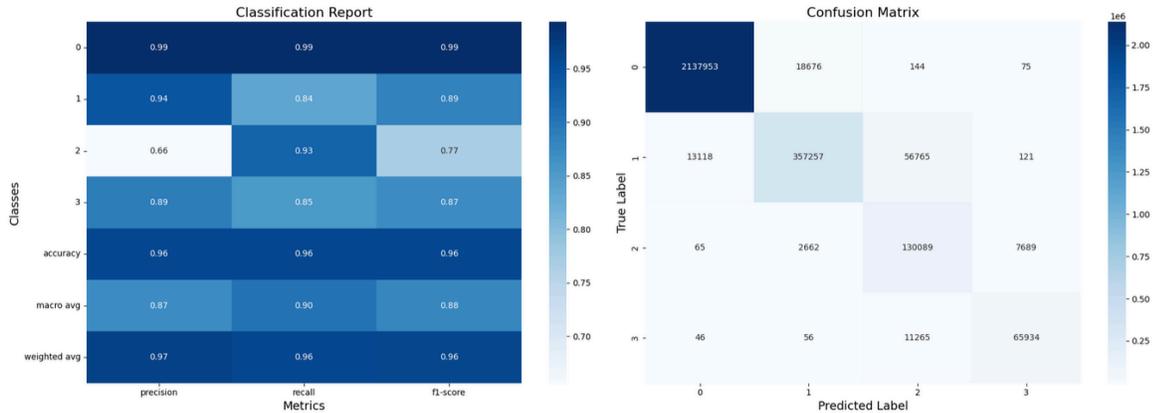

**Figure 12. Confusion matrices and classification reports for LSTM and GRU models for the test set.** The LSTM results are displayed at the top, and the GRU results at the bottom.

### 3.4.3. Comparison between GBMs and Sequential Models

While sequential models showed marginal improvements in some metrics, the gains did not justify the significantly longer training times and increased computational resources.

Concerning the Interpretability, GBMs offer greater transparency through feature importance metrics. For the Training Efficiency, GBMs trained much faster than sequential models (less than 3 minutes vs. 37.5 hours, per model). Finally, for the Performance Trade-offs, Sequential models performed slightly better in macro average F1 scores but were less practical for rapid deployment.

**Performance Metrics:**

The Table 12 provides a performance comparison between GBMs and sequential models. For a fairer assessment, we include the best fine-tuned GBM models optimized via HyperOpt TPE, alongside the performance results of the sequential models.

| Model Configuration | Accuracy | Precision Average | Sensitivity Average | Specificity Average | F1-Score Average | Macro Avg Precision | Macro Avg Sensitivity | Macro Avg F1 | Weighted Avg Precision | Weighted Avg Sensitivity | Weighted Avg F1 | MCC | AUROC Average | AUPRC Average |
|---|---|---|---|---|---|---|---|---|---|---|---|---|---|---|
| XGBoost Opti AUC | 0.95 | 0.85 | 0.9075 | 0.9825 | 0.87 | **0.85** | **0.91** | **0.87** | 0.95 | 0.95 | 0.95 | 0.8851 | 0.995 | 0.9325 |
| CatBoost Opti AUC | **0.96** | 0.85 | 0.9075 | 0.9825 | 0.87 | **0.85** | **0.91** | **0.87** | 0.96 | 0.96 | 0.96 | 0.8894 | 0.995 | 0.9325 |
| LightGBM Opti LogLoss | 0.95 | 0.85 | 0.885 | 0.9525 | 0.87 | **0.85** | **0.91** | **0.87** | 0.96 | 0.96 | 0.96 | 0.8866 | 0.995 | 0.9325 |
| LSTM Model | 0.95 | 0.8661 | 0.9015 | 0.9821 | 0.8768 | **0.85** | **0.91** | **0.87** | 0.96 | 0.95 | 0.96 | 0.8907 | 0.995 | 0.9325 |
| GRU Model | 0.95 | **0.872** | **0.9016** | **0.9853** | **0.8799** | **0.85** | **0.91** | **0.87** | 0.96 | 0.96 | 0.96 | 0.898 | 0.995 | 0.945 |

**Table 12: Performance Comparison of Best Fine-Tuned-GBMs and the Sequential Models.** This table compares performance metrics for the best fine-tuned Gradient Boosting Models (XGBoost Opti AUC, CatBoost Opti AUC and LightGBM Opti LogLoss) and sequential models (LSTM, GRU) across configurations. Indeed, for a fair comparison, GBMs were optimized using HyperOpt TPE with objective functions like AUC and LogLoss, which provided slight performance gains and highlighted the value of hyperparameter tuning in general. Some metrics are shown as averages across all classes, providing a streamlined view (see appendix for detailed class metrics). Both GBM and sequential models consistently score high, demonstrating their effectiveness in multi-class classification tasks. Sequential models also perform well with a 5-minute sliding window approach, showing their suitability for time-series classification in general. Due to long training times, sequential models were not fine-tuned but used complex architectures as proof of concept. Note that decimal precision varies in reporting; for instance, Specificity Average for GRU is 0.9853, though this does not imply any advantage over models with fewer decimals. Both GBMs and sequential models deliver strong, competitive performance in multi-class classification but with pros and cons. See appendix A3 for all models' comparison (including the baseline models).

### 3.5. Feature Importance and Model Interpretability

The feature importance rankings revealed that $SpO_2$ levels, heart rate, age, and respiratory rate were among the most significant predictors, which aligns well with clinical expectations. Notably, this analysis suggests clear patterns between $SpO_2$ and other physiological variables, supporting one of our hypotheses that patterns among features exist and are utilized by the model during training.

In focusing on the most important features for the gradient-boosting models (GBMs), we present the top 25 for clarity and interpretability. Both the XGBoost and LightGBM models underscored the importance of Mask Features, which differentiates between real and synthetic data derived from imputations and later interpolations, corroborating findings from previous studies [42–45]. Additionally, TAG score features consistently ranked quite highly across models, further affirming the utility of the NEWS2+ system in creating impactful, predictive features via its TAG scoring approach.

Although race and gender were less influential overall, they were still present among the top 25 features in some gradient-boosting models (GBMs). Their inclusion underscores the need for further investigation to ensure that any potential biases in model predictions are properly understood and managed.

## 4. Discussion

Our study demonstrates that machine learning models, particularly gradient boosting machines (GBMs), effectively predict hypoxemia severity using physiological parameters, among other factors. The strong performance of GBMs suggests that temporal data may not be essential for this prediction task. This aligns with findings from other studies indicating that deep learning models may not be ideal for tabular data, which, in our case, specifically takes the form of a time series [57].

**Key Findings:**

In comparing GBMs and sequential models, GBMs offer a practical balance of performance, interpretability, and training efficiency, while sequential models yield only slight performance gains at a much higher computational cost. Both types show strong, competitive performance in multi-class classification, though with different strengths and limitations. GBMs, while not designed to capture temporal dependencies, provide competitive results and train quickly, making them highly efficient. In contrast, sequential models, which consider temporal information, take significantly longer to train. For the purposes of this study, they were not fine-tuned due to time constraints, though further tuning might improve their performance.

In terms of feature importance, the inclusion of mask and TAG features notably improved model performance, emphasizing the value of transparent data preprocessing. Additionally, the TAG features derived from the NEWS2+ system demonstrated innovation, proving highly effective in enhancing model accuracy.

Regarding class imbalance, the class-weighted strategy allowed the models to handle the unbalanced dataset effectively, particularly in predicting the most severe hypoxemia cases (label 3). Even though label 1 was more prevalent, the models predicted label 3 more accurately, showcasing the success of the class-weighted approach.

The balance between sensitivity and specificity in model performance is challenging [28], especially for medical event predictions. High specificity with low sensitivity may reduce unnecessary interventions and related costs, as seen with D-dimer testing for venous thromboembolism [58], but could miss critical cases due to missing relevant variables or a limited number of outcome events, making such models more suited as a decision support tool rather than a stand-alone diagnostic tool. Conversely, high sensitivity with low specificity might capture non-specific patterns—such as variables associated with hypoxia but not exclusive to it—or detect subtle changes in nonhypoxic cases, leading to excessive false alerts and limiting clinical usability.

**Limitations and Perspectives:**

The data for this study was sourced from the MIMIC-III and IV databases, which may limit the generalizability of the findings. Our preprocessing involved minute-by-minute interpolation, but it is important to note that the data distribution from MIMIC-III and IV reflects a single hospital with no pediatric data (the minimum age post-preprocessing is 12-17 years). We also stratified patient ages into categories (see Table 4), but due to the dataset's lack of pediatric data and its imbalance across race and ethnicity categories, the model's real deployment potential is limited with respect to these critical demographic factors.

Additionally, while the $SpO_2/FiO_2$ ratio ($FiO_2$ represents the Fraction of Inspired Oxygen in the inhaled gas), is a common indicator of respiratory distress in both clinical and pre-hospital settings (e.g., medical evacuation) [59,60], our study takes a different approach due to limitations in the available data. The MIMIC databases lack sufficient $FiO_2$ measurements, with a high proportion of values missing. Relying on the $SpO_2/FiO_2$ ratio would have introduced excessive uncertainty into the classification; instead, we prioritize $SpO_2$ measurements, which provide a more reliable basis for categorization. Imputing $FiO_2$ data to generate an $SpO_2/FiO_2$ ratio would have likely introduced further biases, especially in instances where both $SpO_2$ and $FiO_2$ values required imputation.

In terms of temporal information, sequential models (e.g., LSTM, GRU) did not demonstrate a significant performance advantage over GBMs, indicating that temporal dependencies may not be essential for predicting hypoxemia severity in this context. Nonetheless, to further validate our study, we should consider using a dataset with naturally regular intervals and fewer missing values, such as HiRID [61], which provides continuous physiological values recorded every two minutes, reducing the need for interpolation and imputation.

**Implications for Practice:**

GBMs can be integrated into systems like the VIMY Multi-System for real-time hypoxemia severity prediction, enabling rapid deployment in (pre)-clinical settings. Understanding the correlation between features and their importance is crucial for both model performance and interpretability. The feature importance metrics provide valuable insights for clinicians, aiding in the understanding of model decisions and improving trust and adoption, which ultimately enhances the interpretability of the system.

Additionally, we have a highly heterogeneous population, which is advantageous as our goal is to predict hypoxia in CBRNE situations in a broadly applicable, non-specific manner. A diverse population supports this aim, increasing the likelihood of developing a model with wide-ranging applicability.

**Future Work:**

Several key areas will be explored to enhance the performance and applicability of the models. First, data integration will focus on incorporating data from multiple hospital databases to improve the model's generalizability across diverse populations. The algorithms will then be tested prospectively on critically ill patients before being deployed in pre-clinical settings. Bias mitigation will ensure that the models remain free from biases related to race and gender, promoting fairness in outcomes. Lastly, real-time implementation will involve developing efficient algorithms suitable for deployment in resource-constrained environments, enabling timely and effective predictions in practical settings.

Fine-tuning the sequential models, despite the training time constraints, could likely bring even minor improvements, which in medical settings can be impactful. We also plan to test the recent xLSTM model, known for its efficiency and lighter architecture compared to other attention-based models, to assess its suitability in this context.

## 5. Conclusions

This study presents a proof of concept for using machine learning models to predict hypoxemia severity in (disaster) triage situations. The GBMs demonstrated robust performance and practicality for deployment within the VIMY Multi-System, while the sequential models, though theoretically advantageous for capturing temporal dependencies, provided marginal gains that did not justify the additional computational demands. Advanced feature engineering was conducted through our newly developed early warning system, NEWS2+, which generated feature scores (TAGs) for six primary physiological parameters. NEWS2+ also enabled the creation of SpO$_2$-based hypoxemia severity scores, adjusted for patient age and disease type, providing novel, clinically-informed labels tailored to our severity prediction task. Future work will emphasize enhancing model generalizability, integrating these models into real-time monitoring systems, and further supporting patient outcomes in emergency medical contexts.

## Supplementary Materials

Figure S1: Inclusion Diagram; Figure S2: Comparison of interpolation methods showing physiological variables' maximum values after interpolation; Figure S3: Demographics of the Study Population; Figure S4: Label distribution of hypoxemia severity scores before and after interpolation; Figure S5: Correlation Matrix of Physiological Variables before imputations and interpolations; Figure S6: PCA Analysis; Figure S7: Overview of Main Preprocessing Steps for Sequential Models with a 5-Minute Sliding Window Illustration for the Initial Minutes of an Admission; Figure S8: Shift-Lag Illustration; Figure S9: Training curves of baseline GBMs models; Figure S10: Feature Importance from the GBMs Models; Figure S11: Training Curves of sequential models over epochs; Figure S12: Confusion matrices and classification reports for LSTM and GRU models for the test set.

## Author Contributions

Conceptualization: S.N., M.A., A.W., P.J., S.B.; Data Curation: S.N., S.B.; Formal Analysis: S.N.; Funding Acquisition: M.A., P.J., S.B.; Investigation: S.N., Y.M., M.A., S.B.; Methodology: S.N., Y.M., M.A.; Project Administration: M.A., P.J., S.B.; Resources: M.A., P.J., S.B.; Software: S.N., Y.M.; Supervision: M.A., A.W., P.J., S.B.; Validation: S.N., Y.M., M.A.; Visualization: S.N., Y.M., M.A., A.W., P.J., S.B.; Writing – Original Draft: S.N., M.A.; Writing – Review & Editing: S.N., M.A., A.W., P.J.
All authors have read and agreed to the published version of the manuscript.

## List of Acronyms

SpO$_2$ : peripheral oxygen saturation.
LSTM: long short-term memory.
PPV: positive predictive value.
SaO2: arterial oxygen saturation.
GBT: gradient boosted tree.
AUROC: area under the receiver operating characteristics.
Lin: linear regression.
LR: logistic regression.
ANN: artificial neural network.
XGB: extreme gradient boosting.
RNN: recurrent neural network.
GBM: gradient boosting model.
PaO$_2$: partial pressure of oxygen.

FiO$_2$: fraction of oxygen in inhaled gas.
NN: neural network.
RF: random forest.
HR: heart rate.
BR: breath rate.
SBP: systolic blood pressure.
DBP: diastolic blood pressure.
AG: anion gap.
CNN: convolutional neural network.
MAE: mean averaged error to the range of values.

# Funding

This research was funded by Ivado's "Scientific in Action" program, through a consortium including the SADC-CDSS Lab of the CHU Sainte Justine Hospital in Montreal, Canada, [applicare.ai](applicare.ai) Solutions inc, and Medint CBRNE Group consulting .

# Institutional Review Board Statement

Not applicable.

# Informed Consent Statement

Not applicable.

# Data Availability Statement

The data supporting reported results were obtained from the MIMIC-III and MIMIC-IV databases, which are publicly available after an approval procedure at https://mimic.physionet.org/.

# Acknowledgments

The authors express their sincere gratitude to the Clinical Decision Support Systems (CDSS) Lab at CHU Sainte-Justine in Montreal, the startup Solutions Applicare.ai, and the MEDINT CBRNE Group consulting for their invaluable support and collaboration. Special thanks go to former students Ons Loukil and Wala Bahri for their contributions to specific coding frameworks and their inspiring work. The authors also thank Professor Ioannis Mitliagkas from Mila/Université de Montréal for his administrative support.

# Conflicts of Interest

The authors declare no conflicts of interest.

# Appendix A

The admissions were selected if they included at least one manually chosen pathology of interest from all possible medical conditions, aiming to have a distribution of patients presenting oxygenation issues in the blood. These patients were more likely to experience hypoxemia episodes, thus making their admissions ideal candidates for labeling.

| Category | Pathologies |
|---|---|
| Infectious Respiratory Diseases | - Pneumonia due to other Streptococcus - Influenza due to identified avian influenza virus with pneumonia - Pneumonia due to parainfluenza virus - Pneumococcal meningitis - Pneumonia in cytomegalic inclusion disease - Pneumonia due to other gram-negative bacteria - Lobar pneumonia, unspecified organism - Pneumocystosis - Abscess of lung with pneumonia - Pneumonia due to Klebsiella pneumoniae - Pneumonia due to anaerobes - Pneumonia due to Methicillin susceptible Staphylococcus aureus - Pneumonia due to Escherichia coli (E. coli) - Pneumonia due to other specified bacteria - Pneumococcal pneumonia (Streptococcus pneumoniae pneumonia) - Viral pneumonia, unspecified - Pneumonia in other systemic mycoses - Pneumonia due to other specified organism - Pneumonia due to mycoplasma pneumoniae - Pneumonia due to Streptococcus pneumoniae - Ventilator associated pneumonia - Pneumonia due to respiratory syncytial virus - Other Staphylococcus pneumonia - Influenza due to other identified influenza virus with other specified pneumonia - Methicillin resistant pneumonia due to Staphylococcus aureus - Bronchitis and pneumonitis due to fumes and vapors - Pneumonia in other infectious diseases classified elsewhere - Pneumonia due to Streptococcus, unspecified - Pneumonia due to Methicillin resistant Staphylococcus aureus - Influenza due to unidentified influenza virus with specified pneumonia - Pneumonia due to Staphylococcus, unspecified - Unspecified bacterial pneumonia - Parainfluenza virus pneumonia - Pneumonia due to Pseudomonas - Acute bronchitis - Bronchopneumonia, unspecified organism - Pneumonia due to Escherichia coli - Methicillin susceptible pneumonia due to Staphylococcus aureus - Pneumonia due to other streptococci - Pneumonia due to adenovirus - Pneumonia due to Gram-negative bacteria - Influenza due to identified novel influenza A virus with pneumonia - Influenza with pneumonia - Pneumonia due to other virus not elsewhere classified - Pneumonia due to Haemophilus influenzae (H. influenzae) - Pneumonia in aspergillosis - Pneumonia due to Legionnaires disease - Pneumonia due to Streptococcus, group A |
| Non-Infectious Respiratory Diseases | - Chronic pulmonary heart disease, unspecified - Respiratory abnormality, unspecified - Extrinsic asthma with (acute) exacerbation - Other emphysema - Pneumonitis due to inhalation of food or vomitus - Other pulmonary insufficiency, not elsewhere classified - Extrinsic asthma, unspecified - Hypoxemia - Other chronic bronchitis - Chronic obstructive asthma, unspecified - Asthma, unspecified type, unspecified - ARDS (Acute Respiratory Distress Syndrome) - Desquamative interstitial pneumonia - Unspecified allergic alveolitis and pneumonitis - Cryptogenic organizing pneumonia - Hypersensitivity pneumonitis due to other organic dusts - Idiopathic non-specific interstitial pneumonitis - Exercise induced bronchospasm - Asthma, unspecified type, with status asthmaticus - Chronic obstructive asthma with status asthmaticus - Asthma, unspecified type, with (acute) exacerbation - Other respiratory complications - Other respiratory abnormalities - Other chronic pulmonary heart diseases |
| Cardiovascular Complications | - Left heart failure - Heart failure, unspecified |
| Hemorrhage and Related Conditions | - Hemorrhage complicating a procedure - Intracerebral hemorrhage - Subarachnoid hemorrhage - Hemorrhage, unspecified - Subdural hemorrhage |
| Poisoning and Toxic Effects | - Poisoning by heroin - Poisoning by cocaine - Poisoning by methadone - Toxic effect of carbon monoxide - Poisoning by other opiates and related narcotics - Poisoning by opium (alkaloids), unspecified - Poisoning by barbiturates |
| Pneumothorax and Related Conditions | - Other spontaneous pneumothorax - Pneumothorax, unspecified - Chronic pneumothorax - Postprocedural pneumothorax - Traumatic pneumohemothorax with open wound into thorax - Traumatic pneumohemothorax without mention of open wound into thorax - Spontaneous tension pneumothorax - Iatrogenic pneumothorax - Traumatic pneumothorax with open wound into thorax - Other pneumothorax |
| Other Respiratory Conditions | - Respiratory distress - Respiratory arrest - Bronchopneumonia, organism unspecified - Postprocedural aspiration pneumonia |

**Appendix A1 Selected Pathologies in the admissions**: the pathologies were associated with low blood oxygenation levels, chosen to reflect the desired patient-type distribution. These remaining pathologies among admissions, manually selected by medical doctors, are represented here in an arbitrary manner for quick reading overview.

The feature columns used to train the Gradient Boosting Machines (GBMs) and sequential models totaled 41 and 45, respectively. However, for the sequential models, additional input columns—'charttime', 'hadm_id', and 'subject_id'—were included. These columns were utilized by the Dataloader to organize the data for sequential processing but were not used directly in training or prediction. Both models used the *hypoxemia_severity_score* column as the target label, and shared a common set of features. This target label was used by the sequential model for the sliding window (5 rows) approach.

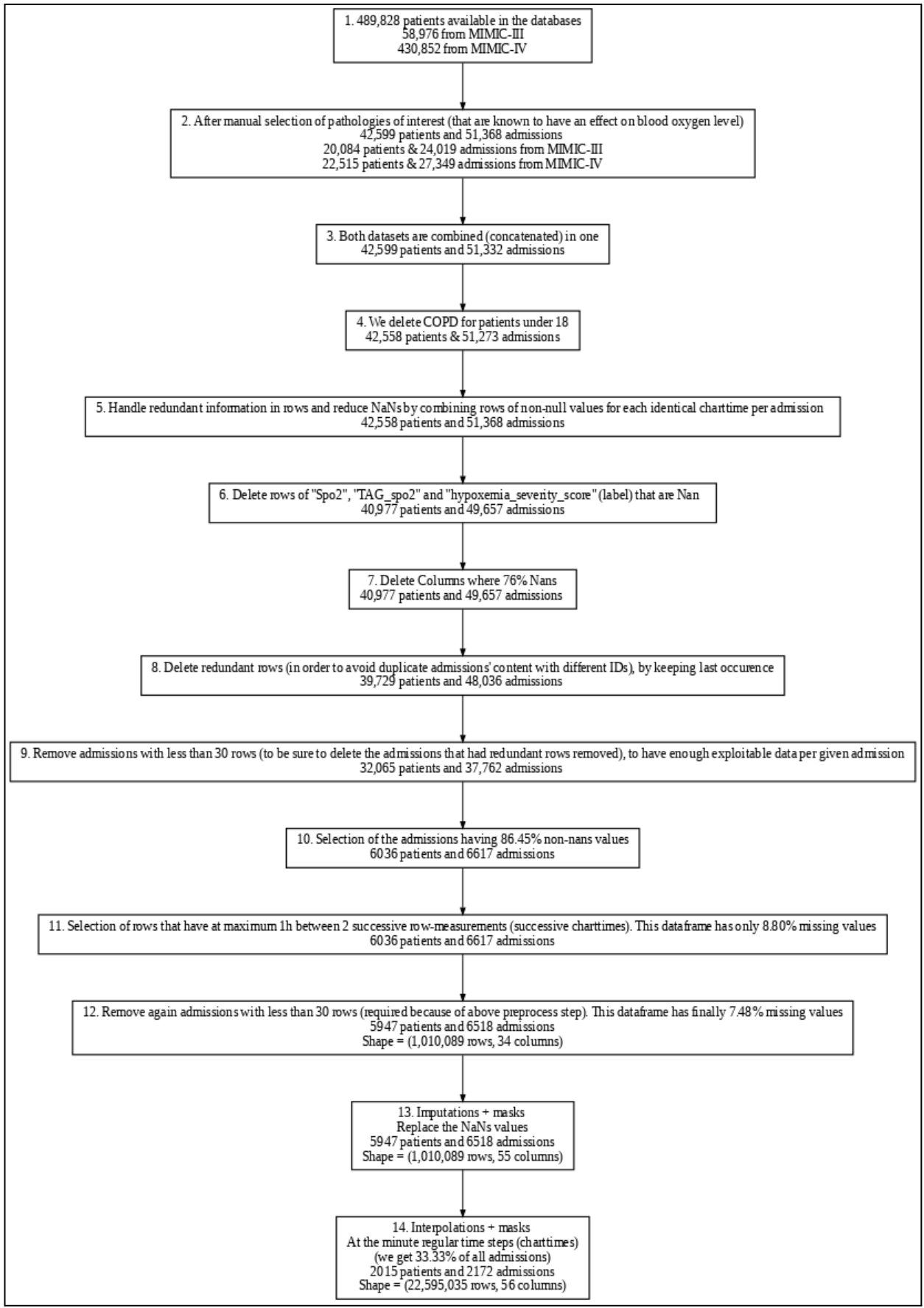

**Appendix A2. Inclusion diagram showing the main preprocessing steps before the imputations and interpolations (until point 12) and after (point 13-14).** In summary: (1) Start with raw data from MIMIC-III and MIMIC-IV; (2) Select pathologies of interest from both dataframes to include patients with blood oxygenation issues; (3) Combine both datasets into one large dataframe; (4) Exclude admissions for patients under 18 years old or suffering from chronic obstructive pulmonary disease (COPD) as the proposesEWS, NEWS2+, does not handle these for labeling; (5) Remove redundant rows (artifacts from previous preprocessing steps) by combining rows with non-null values from identical chart time per admission; (6) Delete rows with NaNs in feature columns of interest ($SpO_2$ linked) to maximize labeled data; (7) Remove rows where 76% or more values are NaNs; (8) To avoid redundant admissions from combined datasets, for the redundant rows that exist, keep the last occurrence of each concerned row and delete earlier ones; (9) Remove rows with fewer than 30 chart times to ensure sufficient data length. This will also lead to the complete deletion of redundant admissions from the previous step; (10) Select admissions with at least 86.45% non-NaN values; (11) Ensure each admission has a maximum of 1 hour between successive rows. Remarkably, this step did not delete any patient or admission from the previous step; (12) Finally, remove admissions with fewer than 30 rows to maintain a certain length; (13) Imputations were applied to replace NaN values with corresponding masks for the affected features, allowing the model to be "informed" about which values are synthetic (i.e., interpolated or imputed) and which are original. This approach ensures the model can distinguish between real and synthetic data during the learning process.; (14) Interpolations **at the minute level** to have regular time steps (in the rows concerned by the charttimes column) with mask for the rows concerned, as well as the features inside it. The resulting dataframe was then used for specific final preprocessing steps tailored for the GBMs and sequential models.

| Model Configuration | Accuracy | Precision class 0 | Precision class 1 | Precision class 2 | Precision class 3 | Sensitivity (Recall) class 0 | Sensitivity (Recall) class 1 | Sensitivity (Recall) class 2 | Sensitivity (Recall) class 3 | Specificity class 0 | Specificity class 1 | Specificity class 2 | Specificity class 3 | F1-Score class 0 | F1-Score class 1 | F1-Score class 2 | F1-Score class 3 | Macro Avg Precision | Macro Avg Sensitivity | Macro Avg F1 | Weighted Avg Precision | Weighted Avg Sensitivity | Weighted Avg F1 | MCC | AUROC class 0 | AUROC class 1 | AUROC class 2 | AUROC class 3 | AUPRC class 0 | AUPRC class 1 | AUPRC class 2 | AUPRC class 3 |
|---|---|---|---|---|---|---|---|---|---|---|---|---|---|---|---|---|---|---|---|---|---|---|---|---|---|---|---|---|---|---|---|---|
| XGBoost Baseline | 0.95 | 0.99 | 0.95 | 0.67 | 0.79 | 0.99 | 0.81 | 0.86 | 0.97 | 0.98 | 0.99 | 0.97 | 0.99 | 0.99 | 0.87 | 0.75 | 0.87 | 0.85 | 0.91 | 0.87 | 0.96 | 0.95 | 0.96 | 0.8882 | 1 | 0.99 | 0.99 | 1 | 1 | 0.93 | 0.85 | 0.95 |
| XGBoost Opti AUC | 0.95 | 0.99 | 0.93 | 0.67 | 0.79 | 0.99 | 0.81 | 0.86 | 0.97 | 0.98 | 0.99 | 0.97 | 0.99 | 0.99 | 0.87 | 0.75 | 0.87 | 0.85 | 0.91 | 0.87 | 0.95 | 0.95 | 0.95 | 0.8851 | 1 | 0.99 | 0.99 | 1 | 1 | 0.93 | 0.85 | 0.96 |
| XGBoost Opti LogLoss | 0.95 | 0.99 | 0.94 | 0.67 | 0.79 | 0.99 | 0.82 | 0.86 | 0.96 | 0.98 | 0.99 | 0.97 | 0.99 | 0.99 | 0.88 | 0.75 | 0.87 | 0.85 | 0.91 | 0.87 | 0.96 | 0.95 | 0.96 | 0.888 | 1 | 0.99 | 0.99 | 1 | 1 | 0.93 | 0.85 | 0.96 |
| CatBoost Baseline | 0.96 | 0.99 | 0.96 | 0.67 | 0.79 | 0.99 | 0.81 | 0.86 | 0.98 | 0.98 | 0.99 | 0.97 | 0.99 | 0.99 | 0.88 | 0.75 | 0.87 | 0.85 | 0.91 | 0.87 | 0.96 | 0.96 | 0.96 | 0.8912 | 1 | 0.98 | 0.99 | 1 | 1 | 0.92 | 0.85 | 0.95 |
| CatBoost Opti AUC | 0.96 | 0.99 | 0.95 | 0.67 | 0.79 | 0.99 | 0.81 | 0.86 | 0.97 | 0.98 | 0.99 | 0.97 | 0.99 | 0.99 | 0.87 | 0.75 | 0.87 | 0.85 | 0.91 | 0.87 | 0.96 | 0.96 | 0.96 | 0.8894 | 1 | 0.99 | 0.99 | 1 | 1 | 0.93 | 0.85 | 0.95 |
| CatBoost Opti LogLoss | 0.95 | 0.99 | 0.94 | 0.67 | 0.79 | 0.99 | 0.81 | 0.86 | 0.97 | 0.98 | 0.99 | 0.97 | 0.96 | 0.99 | 0.87 | 0.75 | 0.87 | 0.85 | 0.91 | 0.87 | 0.96 | 0.96 | 0.96 | 0.8866 | 1 | 0.99 | 0.99 | 1 | 1 | 0.93 | 0.84 | 0.95 |
| LightGBM Baseline | 0.95 | 0.99 | 0.95 | 0.66 | 0.79 | 0.95 | 0.85 | 0.87 | 0.87 | 0.98 | 0.99 | 0.97 | 0.99 | 0.99 | 0.87 | 0.75 | 0.87 | 0.85 | 0.91 | 0.87 | 0.96 | 0.95 | 0.96 | 0.8878 | 1 | 0.99 | 0.99 | 1 | 1 | 0.93 | 0.84 | 0.95 |
| LightGBM Opti AUC | 0.95 | 0.99 | 0.94 | 0.66 | 0.79 | 0.95 | 0.85 | 0.87 | 0.95 | 0.98 | 0.99 | 0.97 | 0.87 | 0.99 | 0.87 | 0.75 | 0.87 | 0.85 | 0.91 | 0.87 | 0.95 | 0.95 | 0.95 | 0.8854 | 1 | 0.99 | 0.99 | 1 | 1 | 0.93 | 0.85 | 0.95 |
| LightGBM Opti LogLoss | 0.95 | 0.99 | 0.94 | 0.67 | 0.79 | 0.95 | 0.85 | 0.87 | 0.87 | 0.98 | 0.99 | 0.97 | 0.87 | 0.99 | 0.87 | 0.75 | 0.87 | 0.85 | 0.91 | 0.87 | 0.96 | 0.96 | 0.96 | 0.8866 | 1 | 0.99 | 0.99 | 1 | 1 | 0.93 | 0.85 | 0.96 |
| LSTM Model | 0.95 | 0.99 | 0.96 | 0.67 | 0.79 | 0.9866 | 0.8371 | 0.9346 | 0.8476 | 0.9824 | 0.9872 | 0.9737 | 0.9972 | 0.9907 | 0.8773 | 0.7682 | 0.8711 | 0.85 | 0.91 | 0.87 | 0.96 | 0.95 | 0.96 | 0.8907 | 1 | 0.99 | 0.99 | 1 | 1 | 0.94 | 0.85 | 0.96 |
| GRU Model | 0.95 | 0.99 | 0.94 | 0.67 | 0.79 | 0.9912 | 0.8352 | 0.9259 | 0.853 | 0.9795 | 0.991 | 0.9744 | 0.9971 | 0.9925 | 0.8866 | 0.768 | 0.8726 | 0.85 | 0.91 | 0.87 | 0.96 | 0.96 | 0.96 | 0.898 | 1 | 0.99 | 0.99 | 1 | 1 | 0.97 | 0.85 | 0.96 |

**Appendix A3:** Detailed comparison table of the performance metrics for Gradient Boosting Models (XGBoost, CatBoost, LightGBM) and Sequential Models (LSTM, GRU) under various configurations. Both sequential models display competitive performance, highlighting their suitability for time-series classification tasks alongside GBMs, though they require significantly longer training times. Optimizing GBMs with HyperOpt via the objective functions (AUC and LogLoss) provides slight improvements, underscoring the impact of hyperparameter tuning on the performance. Note that metrics are reported with varying decimal precision—for example, Specificity for Class 3 in the GRU model appears as 0.9744—though this does not imply an advantage over models with fewer decimal places. Despite these variations, the table offers a comprehensive overview to support analysis and interpretation.